\pdfoutput=1

\documentclass{article}

\PassOptionsToPackage{numbers}{natbib}
\usepackage{amsmath}
    \usepackage[final]{neurips_2021}

\usepackage{subcaption}
\usepackage[utf8]{inputenc} 
\usepackage[T1]{fontenc}    
\usepackage{hyperref}       
\usepackage{url}            
\usepackage{booktabs}       
\usepackage{amsfonts}       
\usepackage{nicefrac}       
\usepackage{microtype}      
\usepackage{xcolor}         
\usepackage{graphicx}
\usepackage{tikz}
\usepackage{comment}
\title{Component Transfer Learning for Deep RL Based on Abstract Representations}

\usepackage{wrapfig}
\usepackage{picins}

\author{%
  Geoffrey van Driessel\\
  VU Amsterdam\\
  \texttt{g.r.van.driessel@vu.nl}\\
   \And
   Vincent Francois-Lavet \\
   VU Amsterdam\\
   \texttt{vincent.francoislavet@vu.nl}\\
   
}

\begin{document}

\maketitle

\begin{abstract}

In this work we investigate a specific transfer learning approach for deep reinforcement learning in the context where the internal dynamics between two tasks are the same but the visual representations differ. 
We learn a low-dimensional encoding of the environment, meant to capture summarizing abstractions, from which the internal dynamics and value functions are learned. Transfer is then obtained by freezing the learned internal dynamics and value functions, thus reusing the shared low-dimensional embedding space. 
When retraining the encoder for transfer, we make several observations: (i) in some cases, there are local minima that have small losses but a mismatching embedding space, resulting in poor task performance and (ii) in the absence of local minima, the output of the encoder converges in our experiments to the same embedding space, which leads to a fast and efficient transfer as compared to learning from scratch.
The local minima are caused by the reduced degree of freedom of the optimization process caused by the frozen models. 
We also find that the transfer performance is heavily reliant on the base model; some base models often result in a successful transfer, whereas other base models often result in a failing transfer.

\end{abstract}


\section{Introduction}
\begin{figure}[htbp]
  \includegraphics[width=1\linewidth]{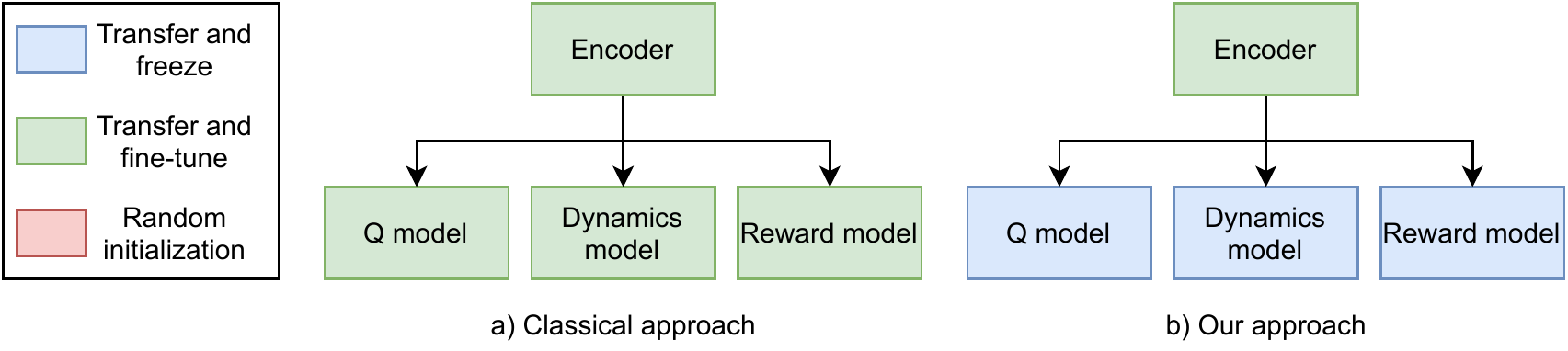}
  \centering
  \caption{Classical fine-tuning approach (a): transfer and fine-tune all models on target task. In comparison, our approach (b) only fine-tunes the encoder by transferring and partially freezing components of model-based RL agent. Note the shared embedding space.}\label{overview}
\end{figure}

State-of-the-art deep reinforcement learning agents are able to beat the best human players in complex games such as Dota 2 \cite{dota2} and StarCraft II \cite{starcraft2}. However training such agents requires many resources and a long training time. For instance the Dota2 agent has been trained using 770±50 PFlops/s·days of compute for 10 months. 
Therefore transfer learning yields an interesting avenue: training a model once to then use it for multiple purposes. 
This is not straightforward because 
a trained agent usually needs to relearn everything from scratch, without gain in performance, when facing an environment with minor visual perturbations such as changing the color of a couple of pixels \cite{image-to-image}.
A popular approach to this problem is fine-tuning (see Figure \ref{overview}a), however some works argue that fine-tuning performs worse than learning from scratch \cite{finetuningfreezing, image-to-image, finetuningresettingconnected, finetuningdifflayers}.


In this work, we leverage the modularity brought by RL techniques based on an abstract representation that is used for both model-free and model-based Deep Reinforcement Learning algorithms (e.g. \cite{ha2018world,CRAR,dreamerv2,muzero}). In that context, only some components are fine-tuned for the new target task. In particular, we consider the visual transfer problems where only the representation of the state space changes such that the encoder has to be fine-tuned and all other components can be frozen (illustrated in Figure \ref{overview}b). We will investigate whether this approach suffers from the same poor fine-tuning performance as reported by the literature. We hypothesize that the frozen models will expect a certain input, and thus enforce an embedding space. More specifically, it will enforce the same embedding space that the agent has found in the source task. We further hypothesize that this method might result in a fast transfer, because 1) only the encoder model weights have to be retrained, and 2) the gradients flowing from the frozen models directly point to an already found global optimum which can help in speeding up the training process.



Our main contributions are the following:
\begin{itemize}
    \item In the context of an RL agent that learns an abstract representation suited for both planning and a value function evaluation, we explore a specific transfer learning procedure. The experimental\footnote{Source code and plot data available at \url{https://github.com/geoffreyvd/deer-clean}
    } results reveal that it works well in many cases but that it can lead into a local minimum other cases.
    \item A study of the local-minimum phenomena suggest that it is caused by 1) small losses area in the optimization landscape at the place of the local minima, and 2) that it gets stuck due to the reduced degree of freedom of the optimization process caused by the frozen models.
\end{itemize}



\section{Related Work}
In the field of transfer learning the most adopted and straightforward approach is fine-tuning. Fine-tuning approaches train an agent on a source task, and then the weights of a number of k layers are used to initialize the weights of a different agent that will be trained on the target task. Besides determining the amount of layers to transfer it can also be decided to freeze the transferred layers. Typically fine-tuning approaches freeze early layers of the convolution network and use a smaller learning rate to fine-tune the rest of the network.

A couple of papers investigated the fine-tuning performance of deep reinforcement learning agents, such as by experimenting with fine-tuning a different amount of layers \cite{finetuningdifflayers}. 
They find that this approach scores better than learning from scratch, suggesting that deep neural networks are able to transfer knowledge. They do not find a major difference in the performance between the amount of layers transferred. 
A similar study \cite{finetuningresettingconnected} finds that fine-tuning all the convolution layers and resetting the fully connected layers has a slightly better performance than learning from scratch.
Two similar studies \cite{finetuningfreezing, image-to-image} reports different findings for experiments in which they freeze the transferred convolution layers. The main findings are 1) freezing layers performs worse than fine-tuning due to negative transfer, 2) learning from scratch generally outperforms the fine-tuning approach and it is more stable, 3) however, fine-tuning can sometimes reach the same or slightly better performance.
This suggests that our approach might also deal with negative transfer due to the frozen layers.


Finally a recent study \cite{fractional} proposed to 1) transfer fractions of the weights and 2) similarly to our work, transfer components of a learned latent dynamics (Dreamer\cite{dreamerv2}). So instead of choosing to transfer or not transfer certain model layers, they introduce a third option which is to fractional transfer the weights of these layers by multiplying the weights with a fraction.
In their experiment they consider a transfer to a target task that has a different reward and action space. And thus they choose to 1) fully transfer the encoder, decoder and RSSM weights, 2) partially transfer action model and 3) fractionally transfer the last layer of the reward and value model. 
The main difference compared to our approach (besides the fractional transfer) is that we freeze all components except the encoder to ideally recover the same embedding space.

\section{Background}
\paragraph{Reinforcement learning}
We define our environment as a Block MDP \cite{blockmdp} defined by the tuple $\langle \mathcal S, \mathcal A, X, p, q, R\rangle$, with a finite
unobservable/hidden state space $\mathcal S$, finite action space $\mathcal A$, finite observation space $X$, latent transition function $p(s'\vert s, a)$ for $s,s' \in \mathcal S$ and $a \in \mathcal A$, emission function q which provides the observation given a latent state $q(h\vert s)$ for $h \in X$ and $s \in \mathcal S$, and finally the reward function $R(s,a,s')$. Then the source and target task have very similar MDP's: $M_{s}$ and $M_{t}$, respectively. Only with different observation space $X_{s}\neq X_{t}$, and emission function $q_{s} \neq q_{t}$. From any state $s$, the objective is to maximize the expected sum of discounted rewards: $\mathbb{E}[\sum_{k=0}^{\infty} \gamma^k R(s_{t+k},a_{t+k},s'_{t+k+1}) \mid s_t=s]$

\paragraph{Transfer learning}
We define transfer learning \cite{ZhuSurvey} as using the source task $M_{s}$ to provide prior knowledge, in our case represented by the transferred model (parameters), with the objective of learning the task in target domain $M_{t}$ better compared to not using prior knowledge.

\section{Method}
Model based methods learn an internal dynamics function to predict what the reward and next (internal) state will be, given a state and action. 
These models often have different losses to train on than the value function. When the encoder is shared and exposed to all losses, the shared embedding space captures features relevant to estimating the value function and the internal model (reward included). Thus we consider model-based DRL agents that utilize a shared encoder prior to the internal dynamics and value function. Meaning that our transfer learning method is applicable to algorithms such as DreamerV2 \cite{dreamerv2} and MuZero \cite{muzero}. Furthermore it could also be applied to supervised or unsupervised deep learning problems for time series.

\subsection{Component Transfer Learning}
Given a fully trained model-based DRL agent on a source task $M_{s}$ (base model), we have access to a solution of the encoder model weights (which outputs an embedding space $x$) and dynamic model weights, for input space $X_{s}$ and transition function $p$ that have small losses. Meaning that the dynamics model is able to estimate correctly the environment transitions given the embedding space $x$. 
When considering a new MDP with a different emission function,
the loss and the gradients will be big again. 
If we freeze the weights of the dynamics model while retraining only the encoder, one strong solution would be that the encoder learns to map the observations (of the new MDP with the new emission function) to exactly the same representation as previously. 
Freezing a good internal dynamics model could thus enforce a certain abstract representation for any MDPs, even with different observations (a different emission function).

\subsection{Description of the different losses}
\begin{figure}[ht!]
 \centering
\scalebox{0.65}{
\begin{tikzpicture}[->,thick]
\small
\tikzstyle{main}=[circle, minimum size = 9mm, thick, draw =black!80, node distance = 12mm]
\tikzstyle{rr}=[minimum size = 7mm, thick, draw =black!80, node distance = 12mm]

\foreach \name in {0,...,2}
    \node[main, fill = red!25] (s\name) at (\name*5,1.5) {$s_\name$};
\foreach \name in {0,...,1}
    \node[rr, fill = red!25] (env\name) at (\name*5+2.5,1.5) {environment};
\foreach \name in {0,...,1}
    \node[main, fill = red!25] (a\name) at (\name*5+1.5,0.5) {$a_\name$};
\foreach \name in {0,...,2}{
    \node[rr, fill = {rgb:black,0.;green,1;blue,1;white,10},text width=1.2cm, align=center] (e\name) at (\name*5+0,0.2) {encoder};%
    }
\foreach \name in {0,...,1}{
    \node[rr, fill = blue!25,text width=2.2cm, align=center, minimum height=6.75em] (mb\name) at (\name*5+2.5,-1.15) {};
    \node[above] at (mb\name.south) {model-based};
    }
\foreach \name in {0,...,1}
    \node[rr, fill = blue!25,text width=1.2cm, align=center] (tr\name) at (\name*5+2.5,-1.4) {transition\\model};
\foreach \name in {0,...,1}
    \node[rr, fill = blue!25,text width=0.9cm, align=center] (rm\name) at (\name*5+2.5,-0.6) {reward\\model};
\foreach \name in {0,...,2}
    \node[main, fill = {rgb:black,0.;green,1;blue,1;white,10},text width=1cm, align=center] (x\name) at (\name*5,-1.2) {abstract\\state};
\foreach \name in {0,...,1}
    \node[main, fill = red!25, align=center] (r\name) at (\name*5+3.5,0.5) {$r_\name$};
\foreach \name in {0,...,2}
    \node[rr, fill = green!40,text width=1cm, align=center] (mf\name) at (\name*5,-2.5) {model-free};
\foreach \name in {0,...,2}
    \node[main, fill = green!40] (V\name) at (\name*5,-3.5) {$Q$};

\node[] (h3) at (3*5-3.3,0) {\huge $\ldots$};

\foreach \h in {0,...,2}
       {
        \path (s\h) edge (e\h);
        \path (e\h) edge (x\h);
        \path (x\h) edge[green!50!black] (mf\h);
        \path (mf\h) edge[green!50!black] (V\h);
        \ifthenelse{\h = 2}{}{
        \path (s\h) edge[red] (env\h);
        \path (a\h) edge[red] (env\h);
        \path (env\h) edge[red] (r\h);
        \path (x\h) edge[blue] (mb\h);
        \path (a\h) edge[blue] (mb\h);
        \path (rm\h) edge[blue] (r\h);
       		}
       }
\foreach \current/\next in {0/1,1/2}
       {
        \path (tr\current) edge[blue] (x\next);
        \path (env\current) edge[red] (s\next);
       }
\end{tikzpicture}
}
\caption{Visualisation of the CRAR architecture \cite{CRAR}. The red components indicate environment attributes. The model-free components are depicted with green, and the model-based with blue. The encoder and thus the abstract state are shared between the two, and depicted with cyan.}\label{CRAR}
\end{figure}
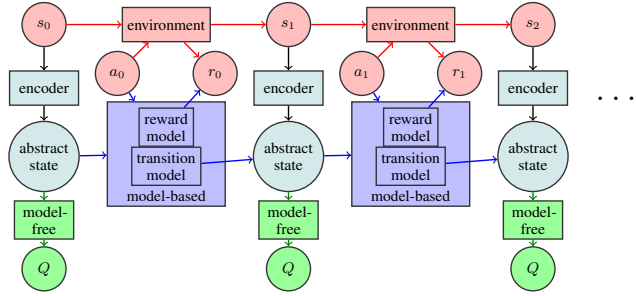

We combine model-free and model-based, which provides the advantage of both approaches; planning and estimation of the cumulative reward. Both are trained on top of a shared low-dimensional embedding space to capture task reward (model-free) and transition (model-based) related information in the same encoding space, as depicted in Figure \ref{CRAR}. Even if the environment has no rewards, the agent is able to create a meaningful embedding space because of a specific loss that prevents the collapse of the representation. 

\paragraph{Notation}
We define the embedding/abstract state space as $x\in \mathcal{X}$ where $\mathcal{X} =\mathbb{R}^{n_m}$ and $n_m\in\mathbb{N}$ is the dimension of the continuous embedding space. Furthermore it defines the following functions: 1) the encoder $\hat{e}:X \rightarrow\mathcal{X}$ is parameterized by $\theta_{\hat{e}}$ which maps the observable state $X$ to the abstract state $x$ 2) the internal transition function $\hat{p}:\mathcal{X} \times A\rightarrow\mathcal{X}$ is paramterized by $\theta \hat{p}$, 3) the reward function $\hat{r}:\mathcal{X}\times A \rightarrow [0,1)$ is parameterized by $\theta_{\hat{r}}$, and 4) the discount function $\hat{\gamma}:\mathcal{X}\times A\rightarrow[0,1)$ parameterized by $\theta_{\hat{\gamma}}$ which is used to predict terminal states.

We use an off-policy learning algorithm that samples transition tuples $(s,a,r,\gamma, s')$ from a replay buffer. Given this tuple the current and next state are encoded using the encoder to generate the embeddings $x\leftarrow\hat{e}(s;\theta_{\hat{e}}), x'\leftarrow\hat{e}(s';\theta_{\hat{e}})$. The Q model is trained using the DDQN algorithm\cite{DDQN} with the target value:
\[Y = r+\gamma Q(\hat{e}(s';\theta_{\hat{e-}}),\operatorname*{argmax}_{a'\in A} Q(x', a';\theta_{Q});\theta_{Q-})\]
Where $\theta_{Q-}$ and $\theta_{e-}$ are the buffered Q and encoder models. Then the agent minimizes the loss:
\[L_{Q}(\theta_{Q}, \theta_{\hat{e}})=(Q(x,a;\theta_{Q})-Y)^2\]
Similar to the CRAR\cite{CRAR} agent, the reward, discount and transition models are trained using:
\[L_{R}(\theta_{\hat{e}},\theta_{\hat{p}})=|r-\hat{r}(x,a;\theta_{\hat{r}})|^2 ,
L_{\gamma}(\theta_{\hat{e}},\theta_{\hat{\gamma}})=|\gamma-\hat{\gamma}(x,a;\theta_{\hat{\gamma}})|^2 ,  L_{p}(\theta_{\hat{e}},\theta_{\hat{p}})=||[x+\hat{p}(x,a;\theta_{\hat{p}})]-x'||^2_{2}\]

Finally we have three embedding structure enforcing losses:
\[L_{d1}(\theta_{\hat{e}})=exp(-C_{d}||e(s_{1};\theta_{\hat{e}})-e(s_{2};\theta_{\hat{e}})||_{2})) 
, L_{d2}(\theta_{\hat{e}})=max(||e(s_{1};\theta_{\hat{e}}||_{\infty}^2)-1,0)\]
From the $L_{d1}$ loss (which forces that two random states have as different embedding values as possible) we introduce the third structure enforcing loss: $L'_{d1}$ which is the same but it samples only successive states. The $L_{d2}$ loss is essentially forcing the embedding space to be in a radius of 1.

\section{Experiments/Results}

\begin{figure}[htbp]
    \begin{minipage}{0.48\textwidth}
        \includegraphics[width=.5\textwidth]{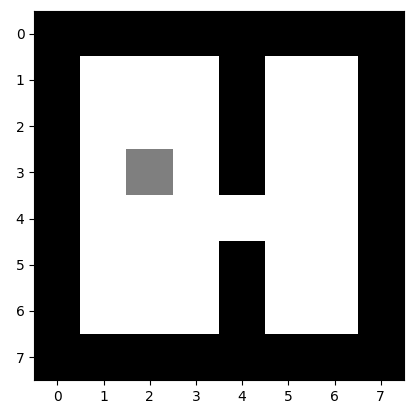}\hfill
        \includegraphics[width=.5\textwidth]{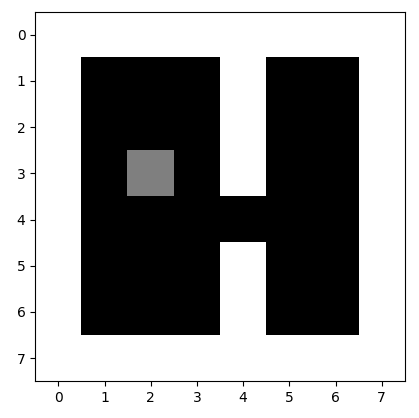}
        \caption{Visualisation of the low-dimensional environment state space, on the right the inverse variant is displayed. }\label{low-dim-maze}
    \end{minipage}\hfill
    \begin{minipage}{0.48\textwidth}
        \includegraphics[width=.5\textwidth]{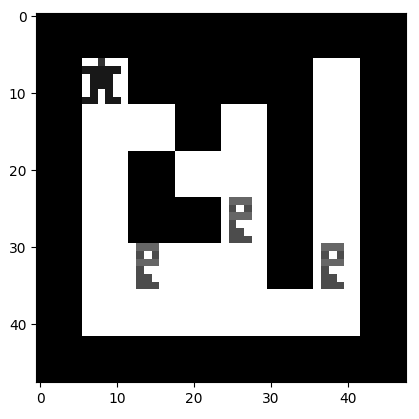}\hfill
        \includegraphics[width=.5\textwidth]{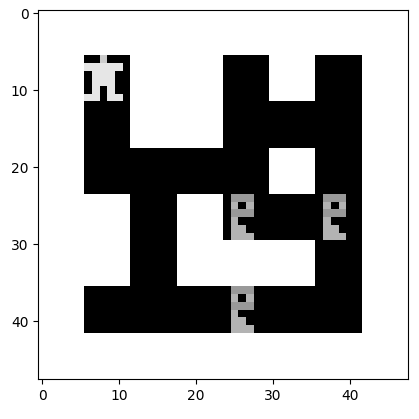}
        \caption{Visualisation of the high dimensional environment state space. Note the random generation of wall, path and reward positions.}\label{highdim_env}
    \end{minipage}
\end{figure}

The section present two experiments: an experiment with a simple maze environment (Figure \ref{low-dim-maze}) and a distribution of mazes (Figure \ref{highdim_env}). 


\subsection{Simple maze}
In this experiment, we consider a simple maze environment without rewards. The environment consists of 8 by 8 tiles, see Figure \ref{low-dim-maze}. Each tile is either a wall, path or the position of the agent. These are represented by a value between 1 and -1. The agent has four possible actions, namely to move in any orthogonal direction. For this environment, the agent is given an embedding space of two continuous variables (two neurons) which allows for simple and direct visualisation and interpretation of the learned embedding space.



The agent is first trained on the regular environment. When it has converged, we use our proposed approach to transfer the learning to the target domain. Meaning that the models weights are used to initialise the models of a second CRAR agent. Then all models are frozen, except for the encoder that is fine-tuned in the target domain. In the transfer phase, the agent is trained on the same maze but with the inverse visuals (target domain), see Figure \ref{low-dim-maze}. 

\subsubsection{Training the base model}
\begin{figure}[htbp]
    \includegraphics[width=.33\textwidth]{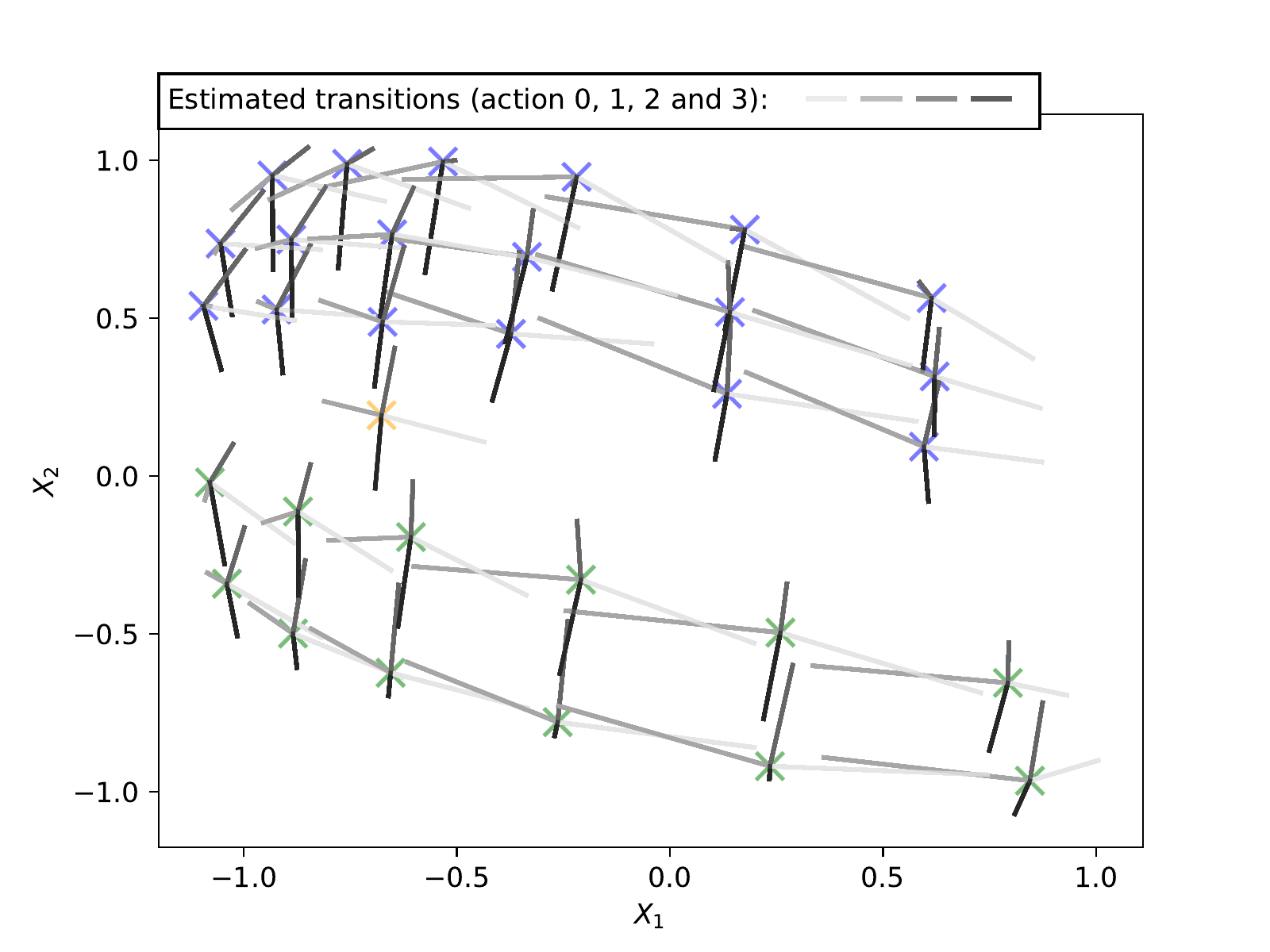}\hfill
    \includegraphics[width=.33\textwidth]{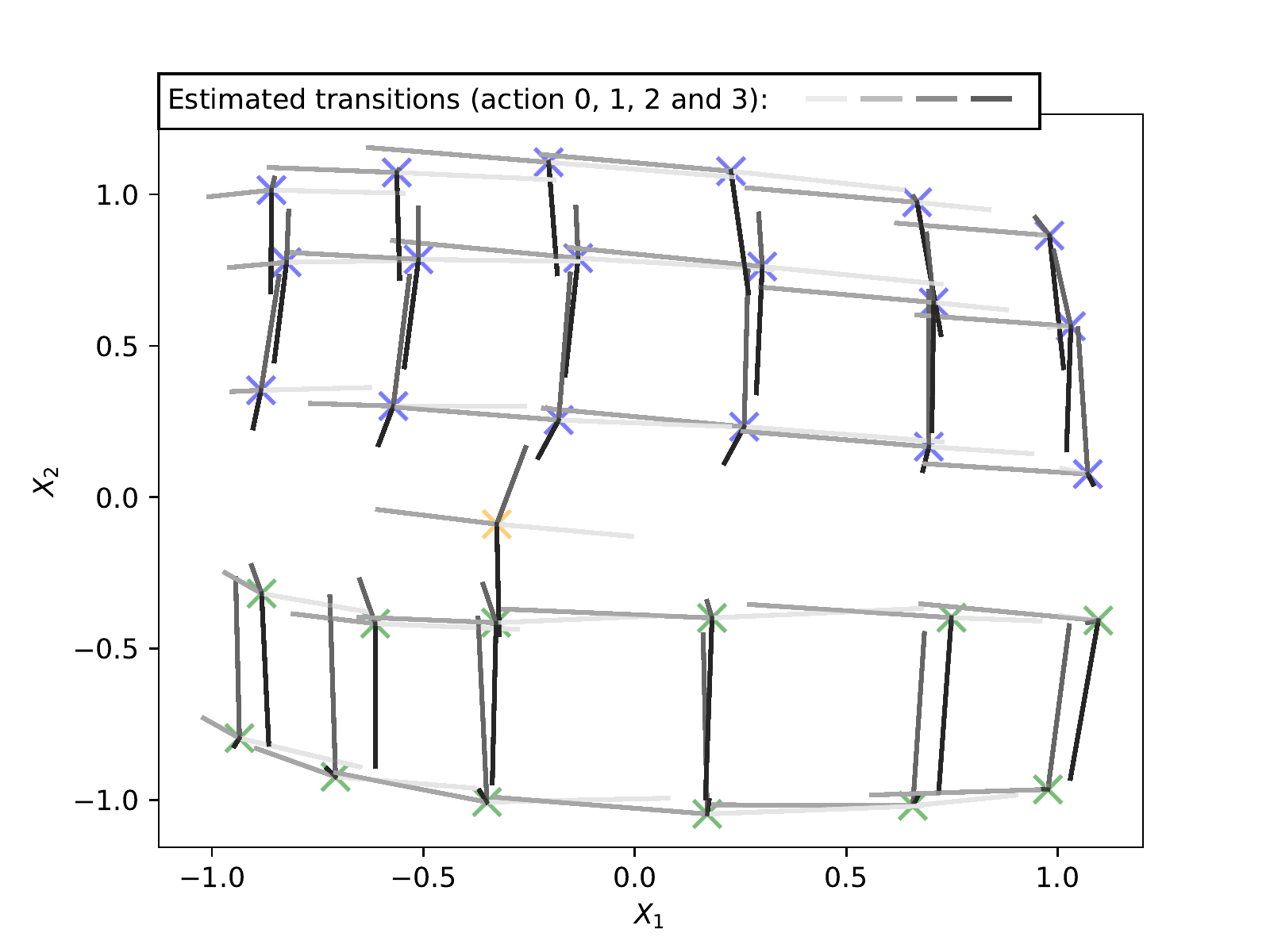}\hfill
    \includegraphics[width=.33\textwidth]{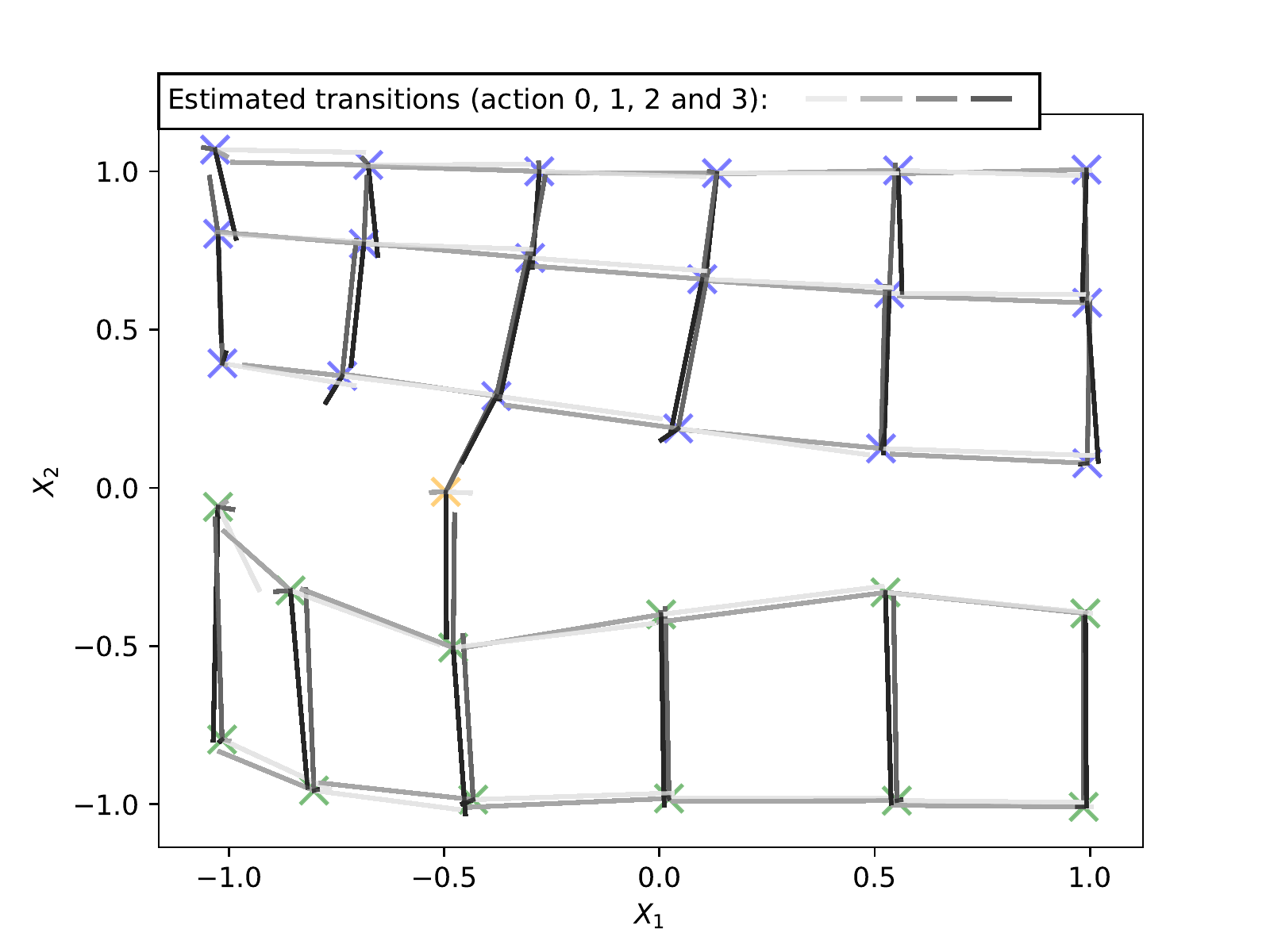}
    \caption{Visualisation of the convergence of the embedding space of a CRAR agent during the training of the simple low-dimensional maze. The crosses indicate the embedded value of a certain state, blue crosses indicate a state of the left side of the maze, and green ones indicate the right side. The lines indicate the corresponding (embedded) transition given an action and a state. From left to right: 5000, 20000 and 160000 (used as base model 1) steps.}\label{base_model}
\end{figure}

We use the default parameters (learning rate = $5 \times 10^{-4}$) to train the base model. From Figure \ref{base_model} we can observe the convergence process of the embedding space. We observe that the left side of the maze is mapped to the top side ($x_{2}>0$) of the embedding space, we thus expect the same in the fine-tuned/transferred embedding space.

\subsubsection{Transferring}
After the base model has been trained, we have executed 30 transfers by fine-tuning the encoder model on the inverse environment (learning rate = $1 \times 10^{-4}$). 
In this simple environment without reward, we can qualitatively asses the performance of the agent by visually inspecting whether the converged embedding space is the same for the base model as for the fine-tuned model.
We find that most transfers successfully converge to the same embedding space, but some fail to converge. 

\begin{figure}[htbp]
    \includegraphics[width=.33\textwidth]{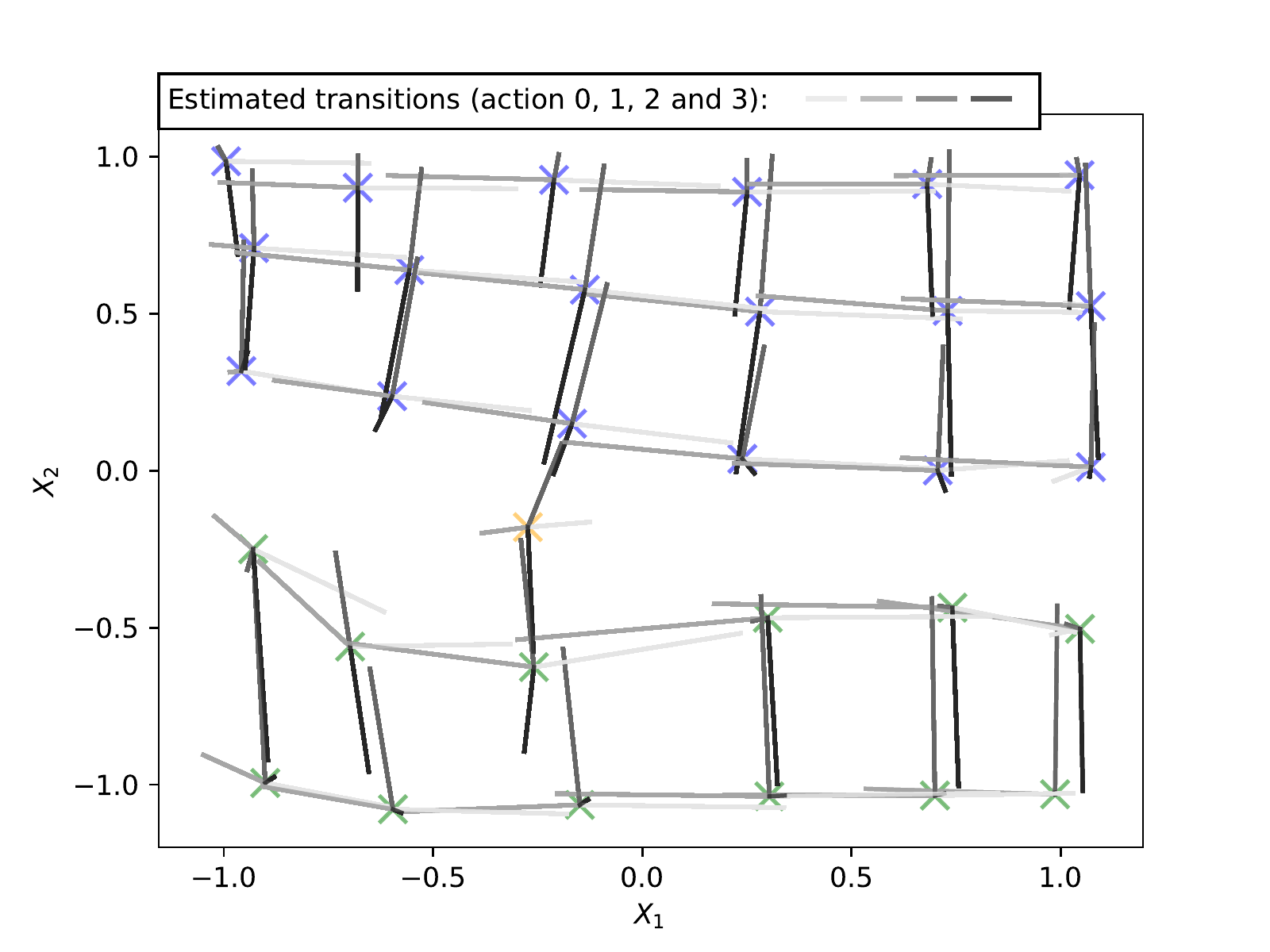}\hfill
    \includegraphics[width=.33\textwidth]{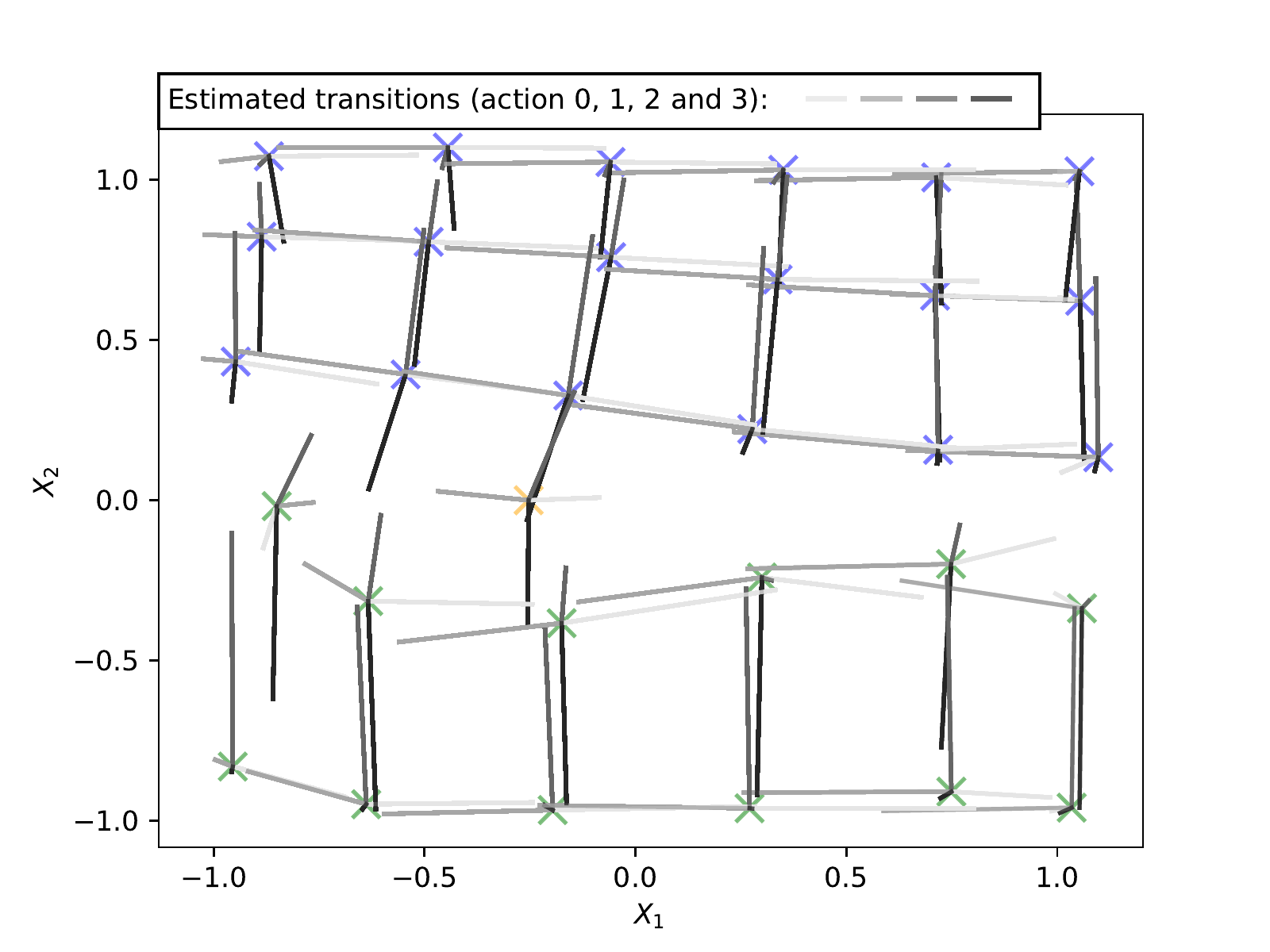}\hfill
    \includegraphics[width=.33\textwidth]{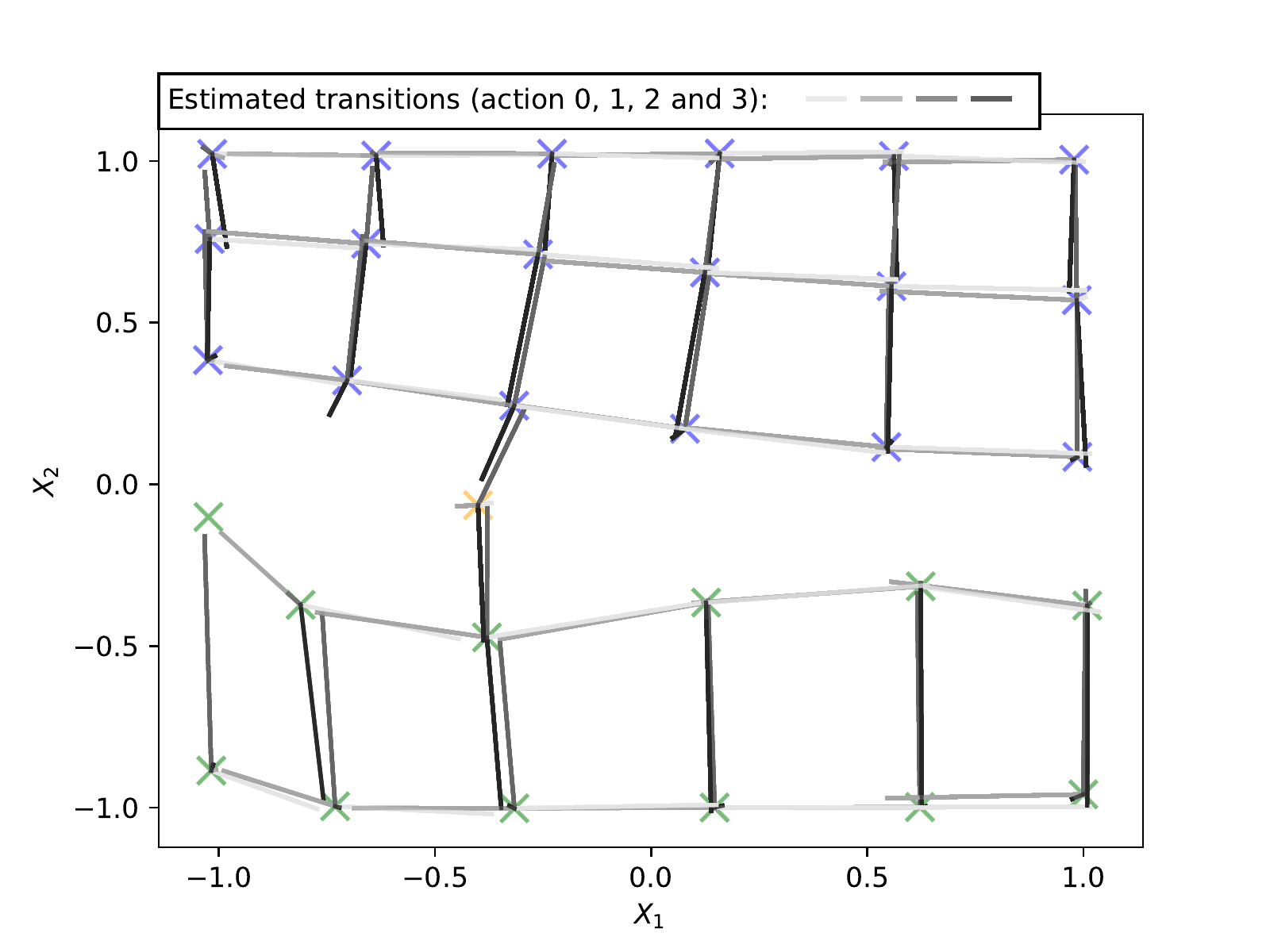}
    \caption{Visualisation of the successful convergence of the embedding space of a CRAR agent during the fine tuning on the inverse low-dimensional maze (base model 1). From left to right: 5000, 20000 and 80000 steps}\label{low_succ}
\end{figure}

\paragraph{Successful transfer}From Figure \ref{low_succ} we can observe that the converged embedding space (step 80000) looks identical to the converged embedding space of the base model (see Figure \ref{base_model}). 
This result looks promising because it appears that the embedding space of the fine-tuning agent converges efficiently to the same representation obtained previously. 

\paragraph{Failing transfer}
Interestingly, we also observe that not all transfers are successful. We trained two CRAR agents on the original environment and used those as base models.
For base model one, 19 out of 22 transfers failed ($90$\%), and for base model two only 1 out of 22 transfers failed ($\sim 5$\%). This might indicate that the quality of the transfer is determined by the frozen models (p-value < 0.0001). 

\begin{figure}[htbp]
    \includegraphics[width=.33\textwidth]{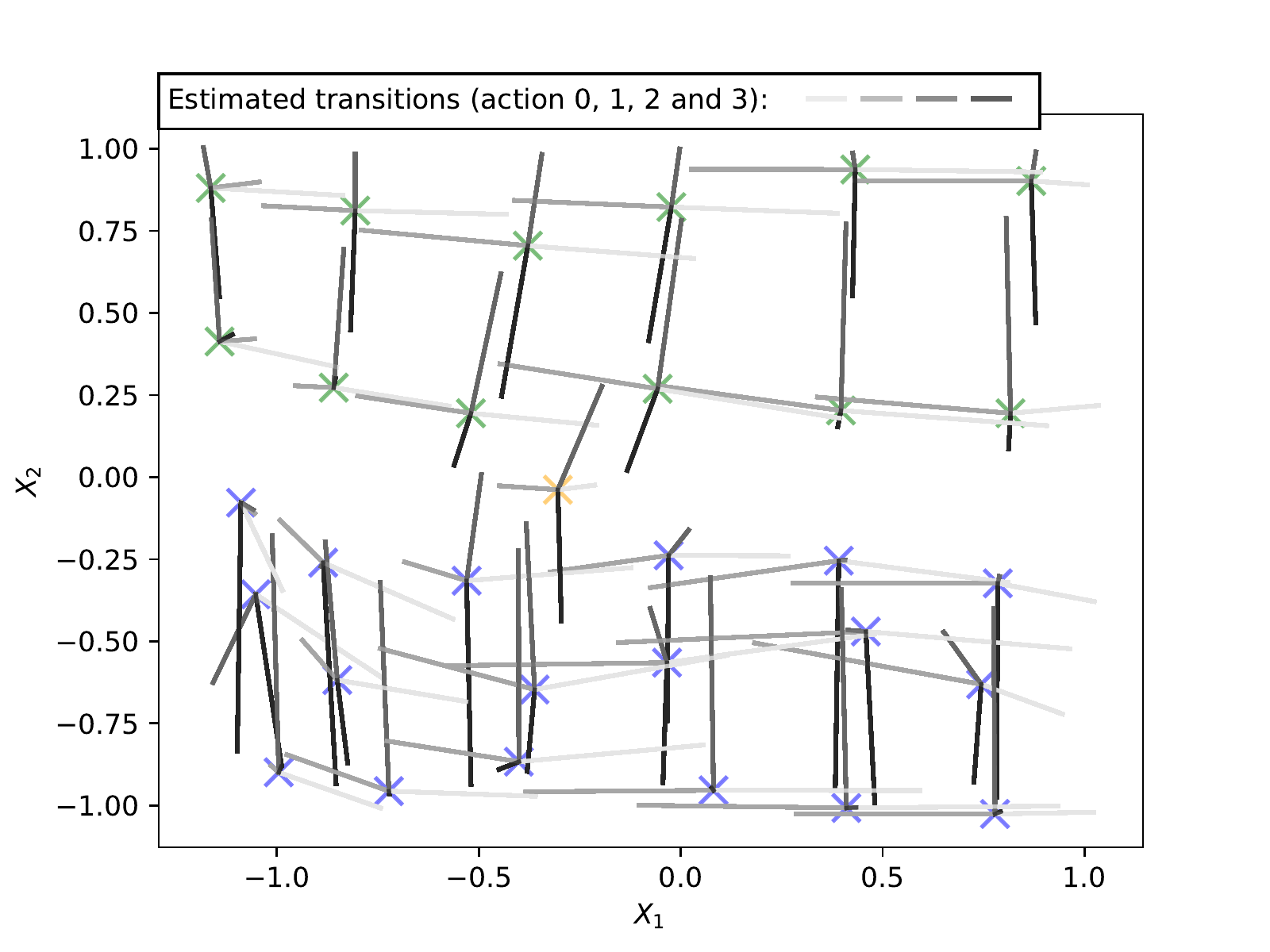}\hfill
    \includegraphics[width=.33\textwidth]{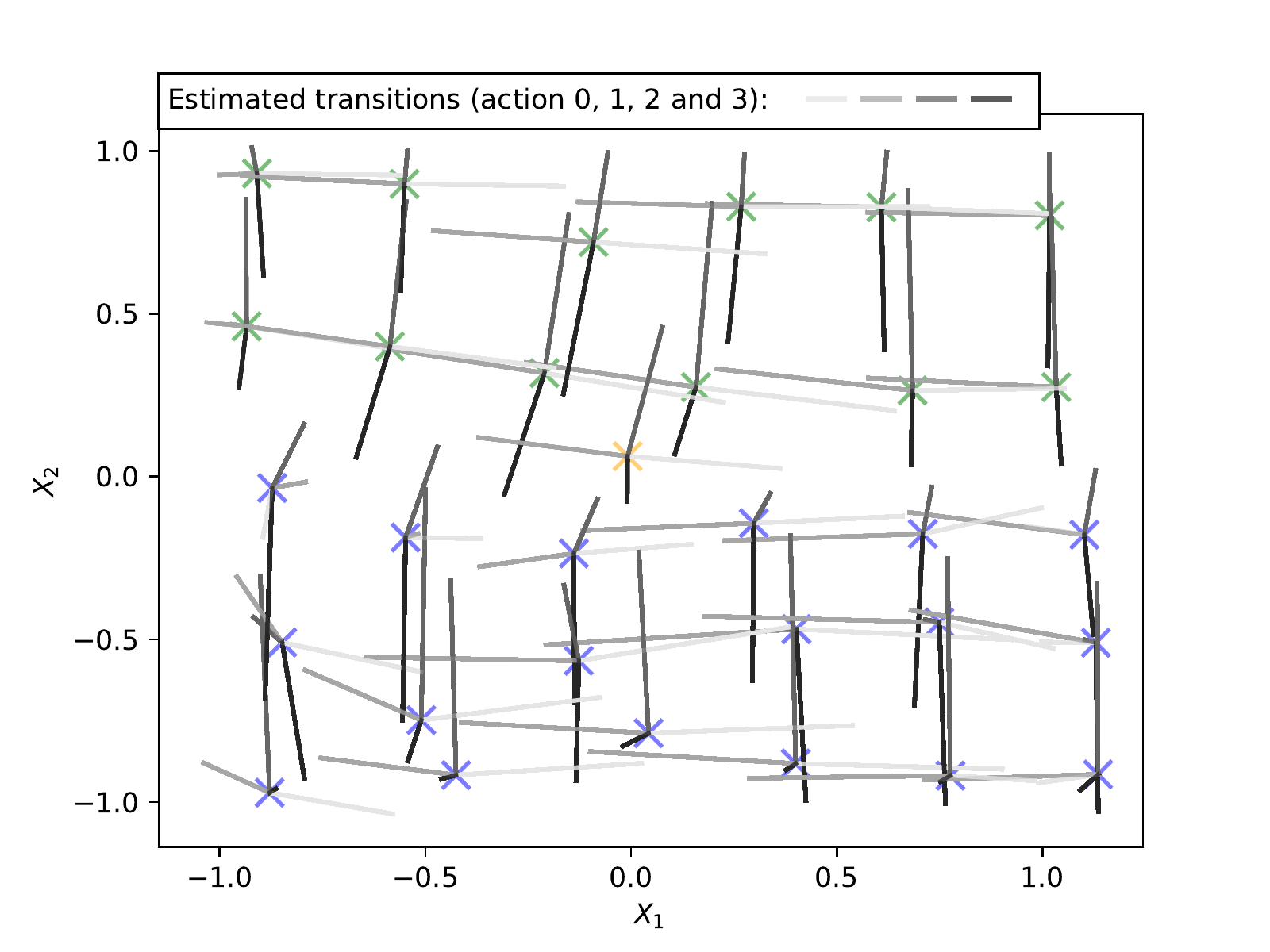}\hfill
    \includegraphics[width=.33\textwidth]{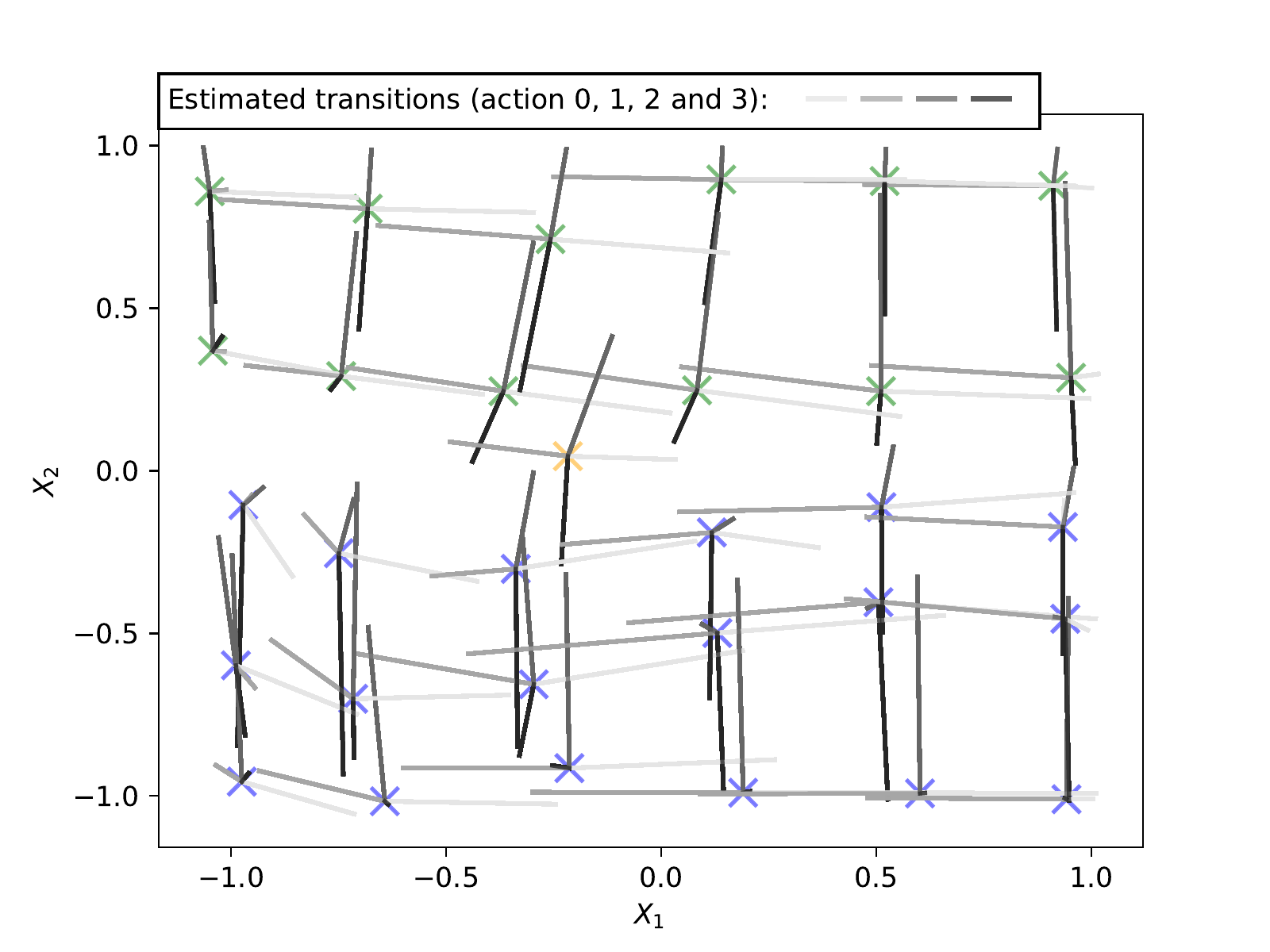}
    \caption{Visualisation of the failing convergence of the embedding space of a CRAR agent during the fine tuning on the inverse low-dimensional maze (base model 1). From left to right: 5000, 20000 and 80000 steps}\label{low_fail}
\end{figure}

From the visualisations of the embedding space of a failed transfer (Figure \ref{low_fail}), we can observe that indeed the agent is not able to converge to the previous found optimum. Furthermore, the structure of the embedding space looks similar to the base model but, if we look at the colors of the crosses, we see that the embedding space has flipped. In this experiment, most of the failing transfers converge to this same embedding space (flipped sides). 

\subsubsection{Interpretation}
We argue that the convergence to the switched mapping is no coincidence. The main argument we make is that by taking the inverse of the environment, it immediately gets caught in quite a ``good" local optimum in terms of the losses but with a very different representation.
An in-depth study of this can be found in Appendix A. 

The previous argument explains why the failed transfers specifically end up in the swapped-embedding local minimum previously visualised. However it only partially explains why it gets stuck in this local minimum (small losses). We argue that the frozen models restrict the degree of freedom in the optimization process. Specifically, we argue that freezing the last layers with respect to the different losses (instead of the encoder losses) changes the neural network degrees of freedom in the optimization process. This in combination with the existence of area with relatively small losses area in the optimization landscape that leads the agent to getting stuck (the maximum entropy losses might also make things worse). See appendix B for a more detailed explanation.

\subsection{High-Dimensional Meta-Labyrinth with rewards}
We now consider a distribution of mazes where each element (wall, agent, reward) are represented as blocks of 6 by 6 pixels, which makes the total input space $48 \times 48$ pixels. In this case, the agent uses an embedding space of 8x8x3 continuous variables and uses convolutional layers in the encoder.
Each maze of the distribution of mazes contains initially three rewards represented by the keys (see Figure \ref{highdim_env}). After collecting the three rewards (or a maximum of 50 steps), the environment generates a new maze in which the positions of the rewards, walls and paths change. Each action that yields to a key leads to a reward of $+1$ and otherwise a reward of $-0.1$. Again we have a normal variant and an inverse variant which will be used as the visual transfer scenario. 


Using this environment we can quantify the performance of a trained model. In this way we will test 1) whether the transfer is successful, 2) what base model has the best transfers, 3) whether the transfer is faster than the training of the base model, 4) compare performance between approaches (e.g. learning from scratch, our approach, some variants of our approach such as resetting the encoder) 5) hyperparameter tuning (learning rate). See Appendix for the latter.

\subsubsection{Impact of base models on transfer performance}
First we trained six base models (basemodel1,...,basemodel6). For each base model, we ran six transfers (using six different seeds) by fine-tuning the encoder on the inverse variant of the environment. This results in a total of 36 transferred agents. The base models will be trained for 200 epochs, and the transfers for 100 epochs. Lastly the learning rate used for the fine-tuning is the same as the training of the base model: $5 \times 10^{-4}$. 

\begin{figure}[htbp]
    \includegraphics[width=.5\textwidth]{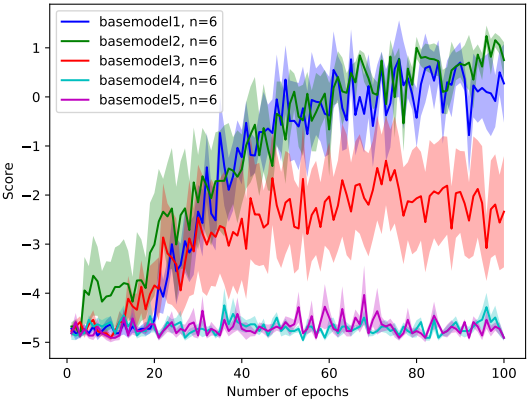}\hfill
    \includegraphics[width=.5\textwidth]{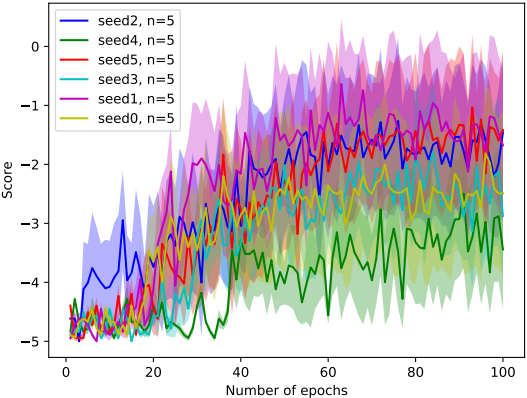}
    \caption{Average and standard error of the performance of the transfers, six transfers per base model (first base model is called basemodel1, etc..). Both plots are generated from the same data but show the average and standard deviation of different groupings, the left plot has the data grouped by the base model, and the right plot has the data grouped by the seed number of the fine tuning process.}\label{highdim_perforamncetrasnfer}
\end{figure}

In Figure \ref{highdim_perforamncetrasnfer} the performance of the transfers can be observed. From the left image we can make some quick conclusions, 1) the variance in the performance is big among different transfers with different base models, but can also be big among the same base model (see performance of group basemodel3), 2) some base models do not have a single succeeding transfer (basemodel4 and basemodel5). 3) some base models appear to always have a successful transfer (basemodel1 and basemodel2). Following these findings we can confirm the same finding as before: the base model that will be fine-tuned on determines the transfer success rate. Meaning that the frozen models determine whether the flowing gradients steer into a local minimum or not. From the image on the right we find what we expect: approximately the same average and standard deviation for the different seeds.

\subsubsection{Study of high-dimensional embeddings}
In this subsection we will look into the difference between the embeddings of successful and failing transfers. This is done 1) quantitatively by calculating the L2 distance between the base model and the transfer and 2) qualitatively by visually inspecting the embedding space. The L2 distance is measured by taking an instance of the environment, and having this same instance rendered in the normal visual variant and the inverse variant. Then the normal visual variant is fed to a base model, and the inverse variant is fed to one of its transferred models. Then the L2 distance is calculated between the embedding of the base model and the embedding of the transferred model.

\begin{figure}[htbp]
    \includegraphics[width=.2\textwidth]{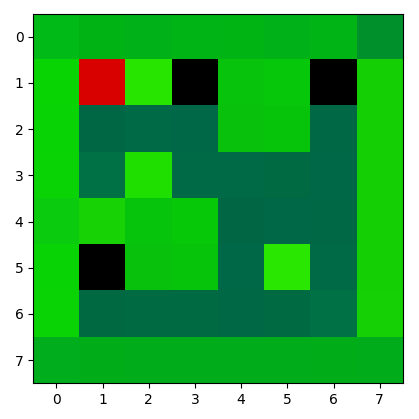}\hfill
    \includegraphics[width=.2\textwidth]{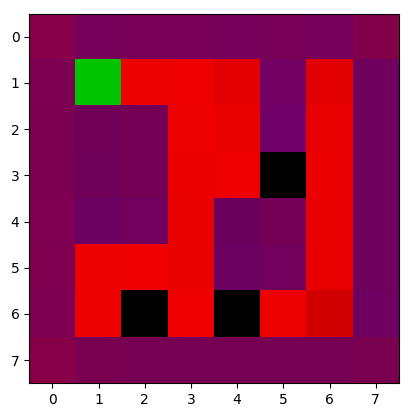}\hfill
    \includegraphics[width=.2\textwidth]{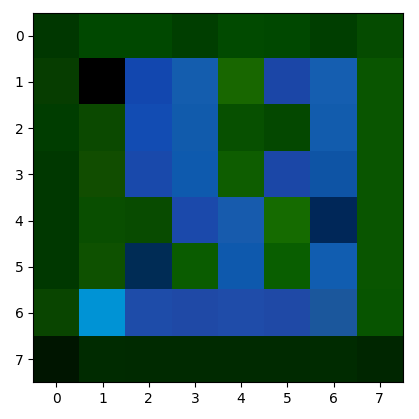}\hfill
    \includegraphics[width=.2\textwidth]{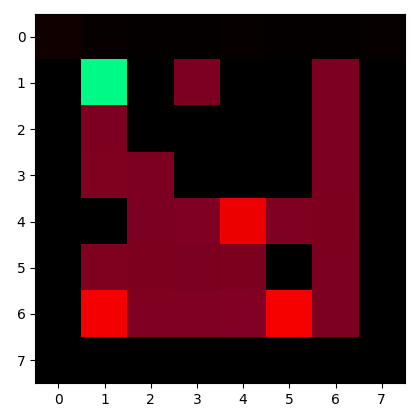}\hfill
    \includegraphics[width=.2\textwidth]{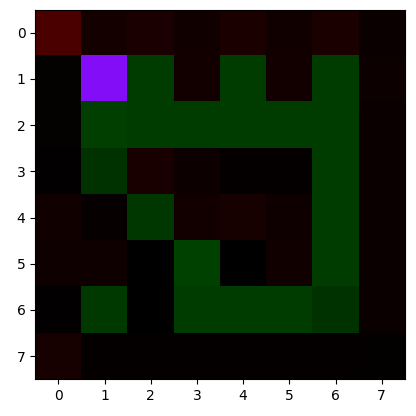}
    \\[\smallskipamount]
    \includegraphics[width=.2\textwidth]{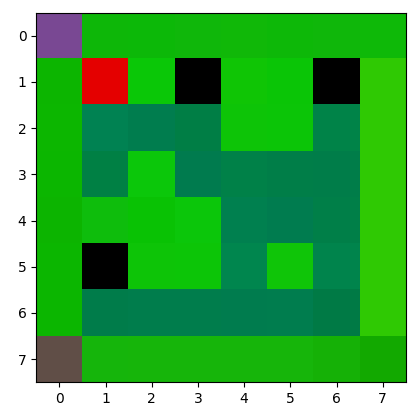}\hfill
    \includegraphics[width=.2\textwidth]{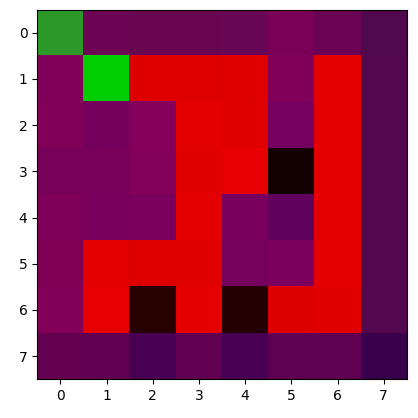}\hfill
    \includegraphics[width=.2\textwidth]{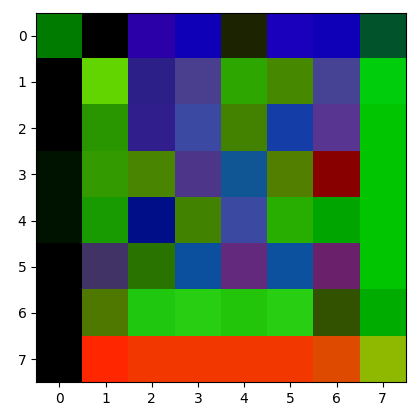}\hfill
    \includegraphics[width=.2\textwidth]{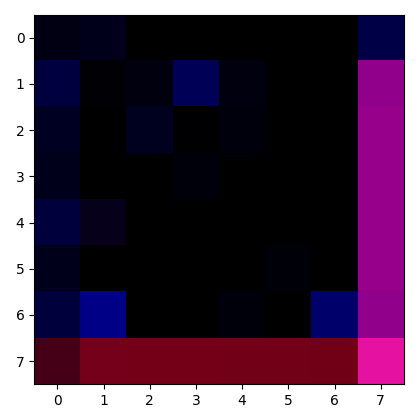}\hfill
    \includegraphics[width=.2\textwidth]{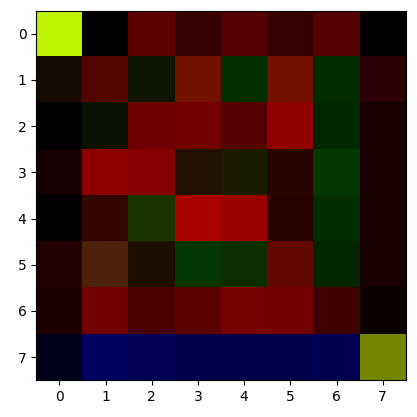}
    \caption{Upper row are embeddings of base models, lower row are embeddings of fine-tuned models (on the inverse environment instance). Each column indicates a different model. So the two figures in the first column, are 1) the original model: basemodel1 and 2) a fine-tuned model on basemodel1. The second column contains the base model basemodel2 and the model fine-tuned on basemodel2, etc. }\label{highdimembedding}
\end{figure}

Measuring the L2 distance reveals that the distance between the transfers and the base model is bimodal, with successful transfers that have a loss around ~1.1 while it is around ~6.2 for the failing transfers. The embedding space of the successful transfer converge close to the embedding space of the base model. This partially confirms our hypothesis: the gradients flowing from the frozen models can stimulate the encoder to converge to the same output space (embeddings)  for visually different inputs but the risk of local minima is also confirmed. An overview of the visualisations of the embeddings of the five base models with an associated transfer is given in Figure \ref{highdimembedding}. We can for instance observe that 1) the embedding of the good base model (basemodel1) is the same as the transfer, and 2) the embedding of the poor base model(basemodel5) has a mismatching embedding with its transfer.


\subsubsection{Performance comparison among approaches}
\begin{figure}[htbp!]
    \includegraphics[width=.5\textwidth]{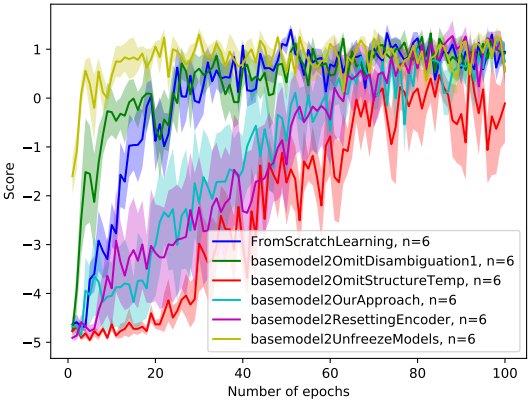}\hfill
    \includegraphics[width=.5\textwidth]{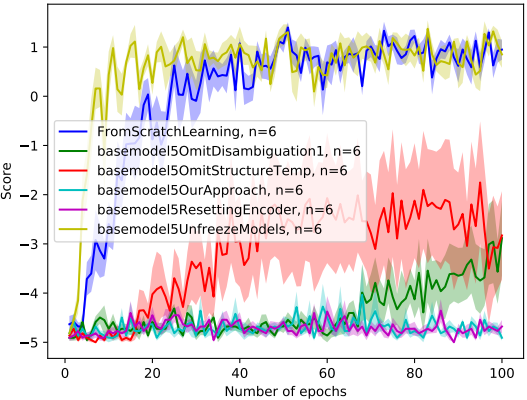}
    \caption{Visualisation of the average and standard error of the score for six approaches. The left depicts the transfer performance of the approaches using the good transfer performing base model (basemodel2), and on the right using the poor transfer performing base model (basemodel5). Note that the spikes in all lines can be explained due to the variance in the difficulty of randomly generated maps; completing the map successfully (collecting three rewards) will yield different scores based on the required amount of steps that have to be taken. 
    }\label{highidim_comparisondiffmethods}
\end{figure}

In this subsection, the performance of four variants of our approach are reported. The first variant is the same as our approach in all regards except that it resets the encoder weights instead of initializing them with the encoder weights of the base model (Figure \ref{var1}). The second variant simply fine-tunes the whole base model on the target domain, so not freezing any model (Figure \ref{overview}a). The third variant again is the same as our proposed approach with the exception that it omits the Disambiguation1 loss. The fourth variant temporarily (for 15 epochs) omits all structure enforcing losses. We evaluate the performance using two different base models, one which had good transfer performance using our original approach, 
and one with poor transfer performance. 
See Figure \ref{highidim_comparisondiffmethods} for an overview of the performance of six approaches: our original approach, the four variants and learning from scratch.


The results of the four variations in the transfer procedure are shown in Figure \ref{highidim_comparisondiffmethods}:
1) resetting the encoder (cyan) helps to avoid a local minimum for transfers (but not in all cases), 
2) fine-tuning all models (purple) performs the best, 
3) omitting the disambiguation1 loss (blue) performs on average better than learning from scratch, 
4) temporarily omitting all structure enforcing losses (green)is the only approach (without fine-tuning) that is able to perform some successful transfers in all cases,
5) from scratch learning (yellow) has a low variance and scores second best, 
and finally 6) our original approach (red) performs the same as resetting the encoder variation. 
This implies that 1) fine-tuning all models indeed resolves the restricted degree of freedom in the optimization process, and 2) changing the loss function in the fine-tuning steps looks promising (interestingly enough it is still able to converge to the same solution although the loss function changed).



\section{Discussion and Conclusion}
When only retraining the encoder, transfer can converge, in some cases, to the same embedding space as the base model. However we also found that our proposed approach does not always converge to the same (optimal) embedding space as the base model, instead getting stuck in a local minimum. Furthermore we find that the success rate of the transfer is heavily dependant on the base model. 
A study of the local minima suggested that they are caused by two problems, namely 1) small losses area in the optimization landscape at the place of the local minima, and 2) the reduced degree of freedom of the optimization process caused by the frozen parts in the neural network models. 

Quantitative results show that 1) learning from scratch is more stable and performs better than retraining only the encoder, and 2) fine-tuning all parts of the model (complete fine-tuning) outperforms all other approaches. 

\bibliographystyle{IEEEtran}
\bibliography{refs} 

\newpage

\appendix

\section{Analysis local minimum}
In the following sections we provide arguments of 1) why the failing transfers of the low-dimensional maze specifically end up in the swapped embedding space local minimum previously visualised and 2) why it gets stuck in this local minimum. 

\subsection{Swapped embedding local minimum}
\begin{figure}[htbp]
    \includegraphics[width=.5\textwidth]{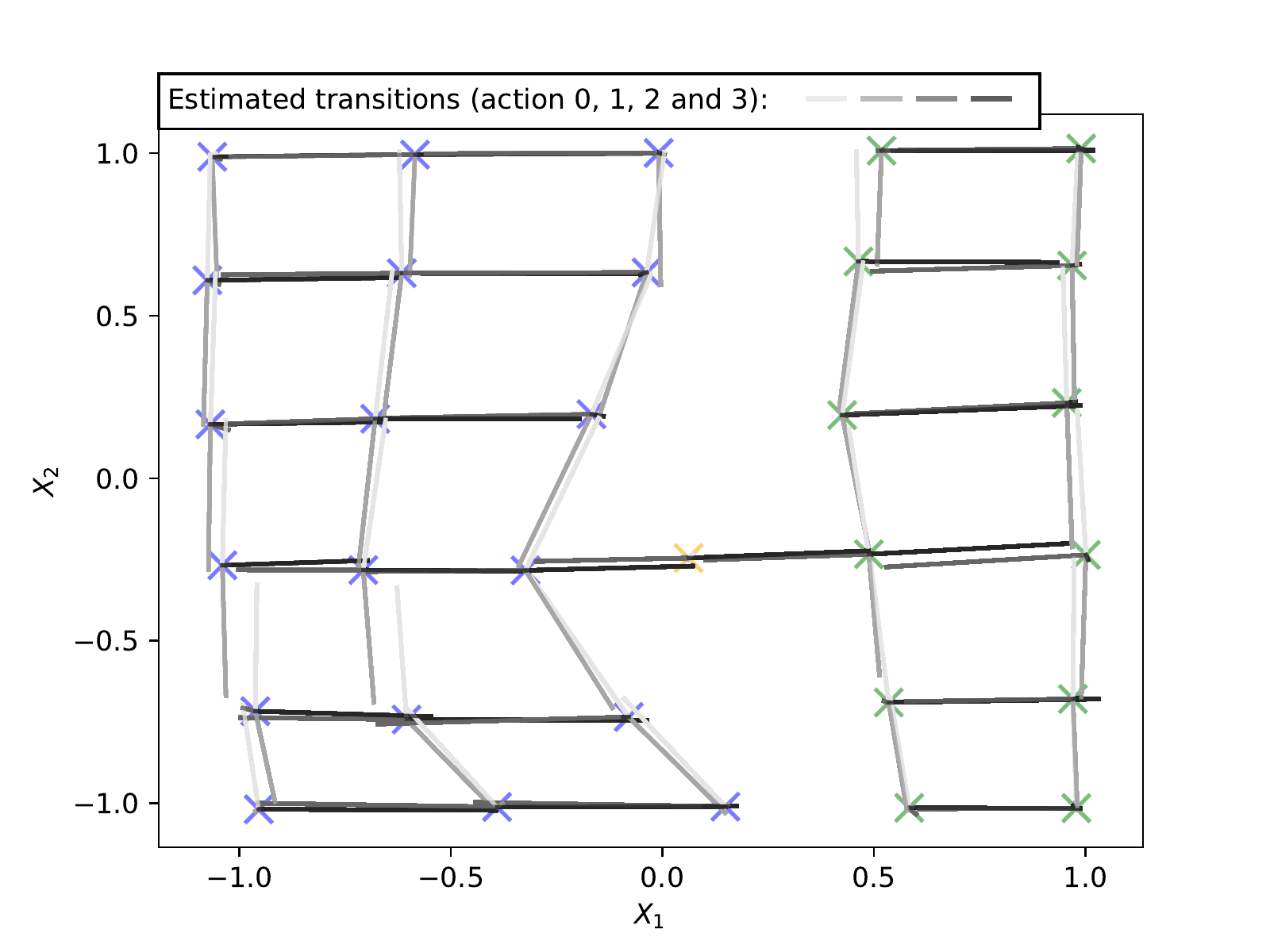}\hfill
    \includegraphics[width=.5\textwidth]{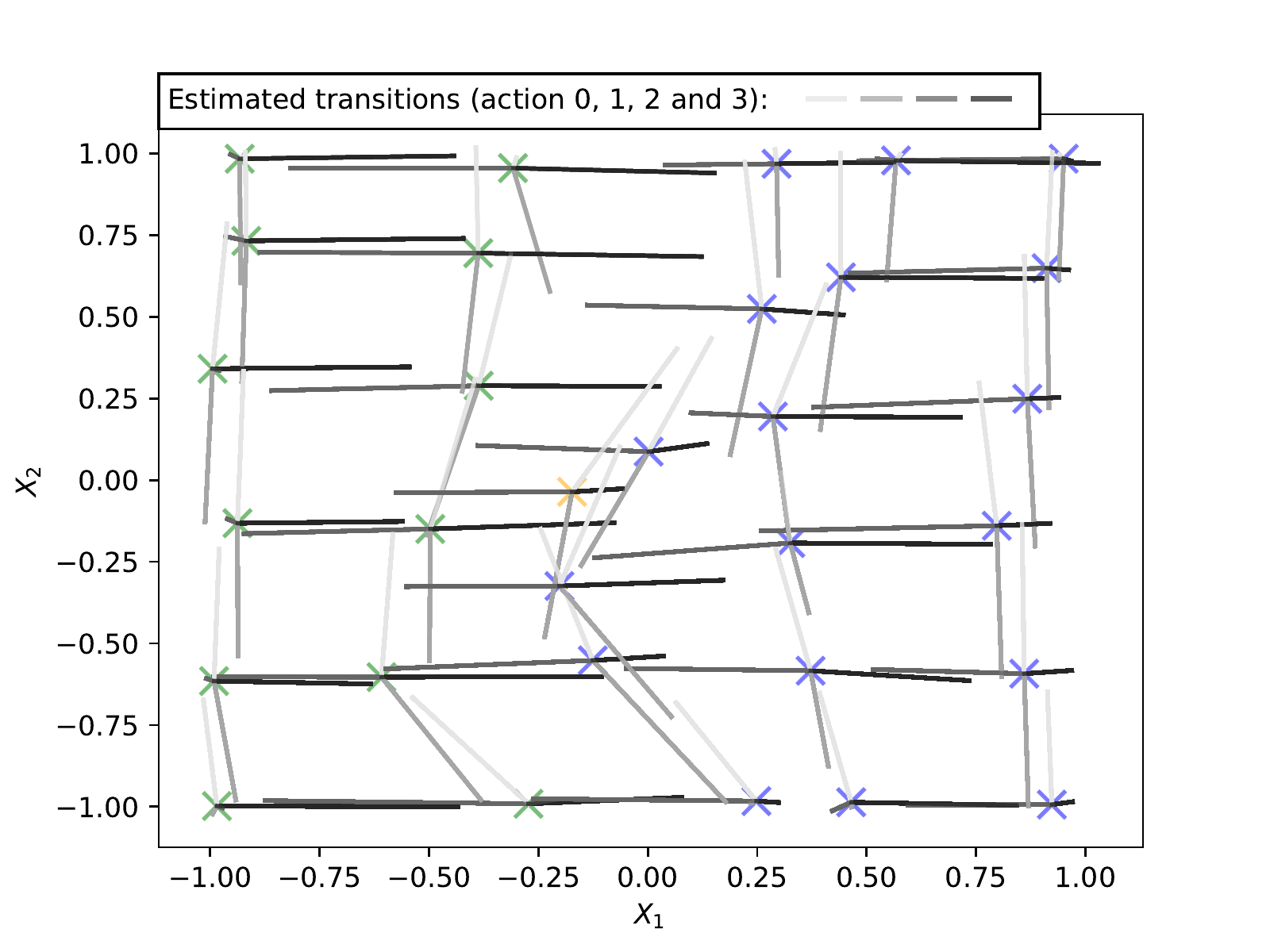}
    \caption{Visualisation of the embedding space learned by the CRAR agent in the low dimensional environment without any rewards (base model 2). The image on the left is the original model learned from scratch, whereas the image on the right is the (failed) fined-tuned model trained in the inverse environment.}\label{low_comparison_embed}
\end{figure}
\begin{figure}[htbp]
    \centering
    \includegraphics[width=0.62\textwidth]{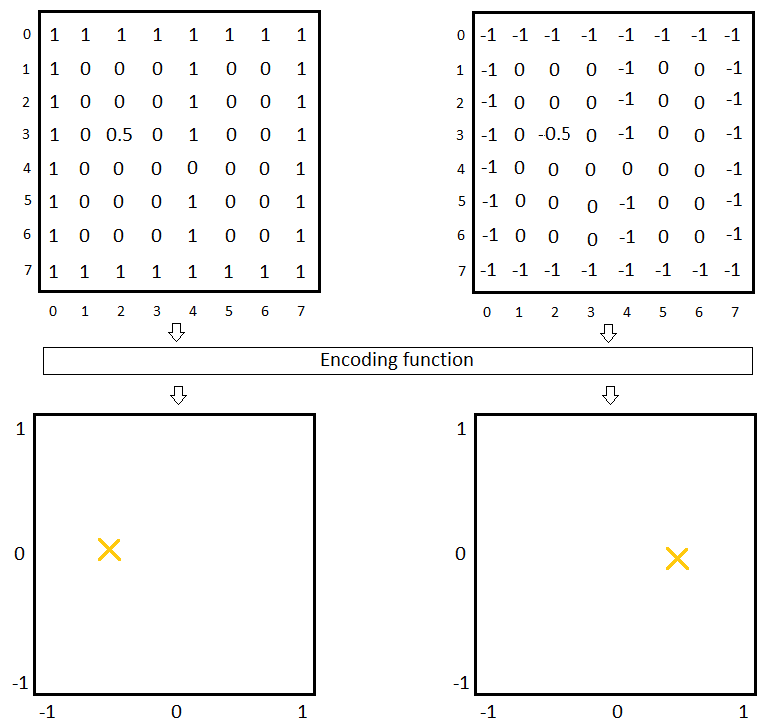}
    \caption{Visualisation of the swapped embedding space phenomena. The first row contains the matrices of the actual input values, for the original  (left) and inverse (right) environment (as visualised in Figure \ref{low-dim-maze}). The second row contains the resulting embedding associated to that input when passed through the simple encoding function.}\label{matrixandlocationinembedding}
\end{figure}

 When we look at the embedding space of the converged base model in Figure \ref{low_comparison_embed} and look at the environment and the position of the agent in Figure \ref{low-dim-maze}, we can conclude that the state space with the agent on position (2,3) is mapped to the embedding value: ~(-0.6, 0.2). For the sake of keeping the example as simple as possible we will round these value to (-0.5, 0). Meaning that the agent has an encoding function that converts the state (which is a 8 by 8 matrix where, for the non-inverse variant, an element has the value 1, 0 or 0.5 indicating a wall, path or player position, respectively) with the player position at (2,3), to the embedding value: (-0.5, 0). This is visualised on the left side of Figure \ref{matrixandlocationinembedding}. 

A solution of such an encoding function could simply be: embeddingX=state(2,3)*-1, embeddingY=0, which results in: -0.5=0.5*-1, 0=0. Now given this function we can feed the inverse state to this function and see what happens. Remember that the inverse environment simply replaced the 1, 0 and 0.5 tile values by -1, 0, and -0.5. Now if we take the same positions of all tiles but in the inverse fashion, such as depicted by the right image of Figure \ref{low-dim-maze} and feed it to the encoding function just defined. We get the following result 0.5=-0.5*-1, 0=0. Meaning that we previously had an embedding of (-0.5, 0) and now have an embedding of (0.5,0) for the same hidden state namely having a state with the player at position (2,3). Which indeed results in the switched sides in the embeddings previously seen. This ofcourse assumes that the simple function can be extrapolated to other states with different player locations as well, which is quite trivial. 

\begin{figure}[htbp]
    \includegraphics[width=.5\textwidth]{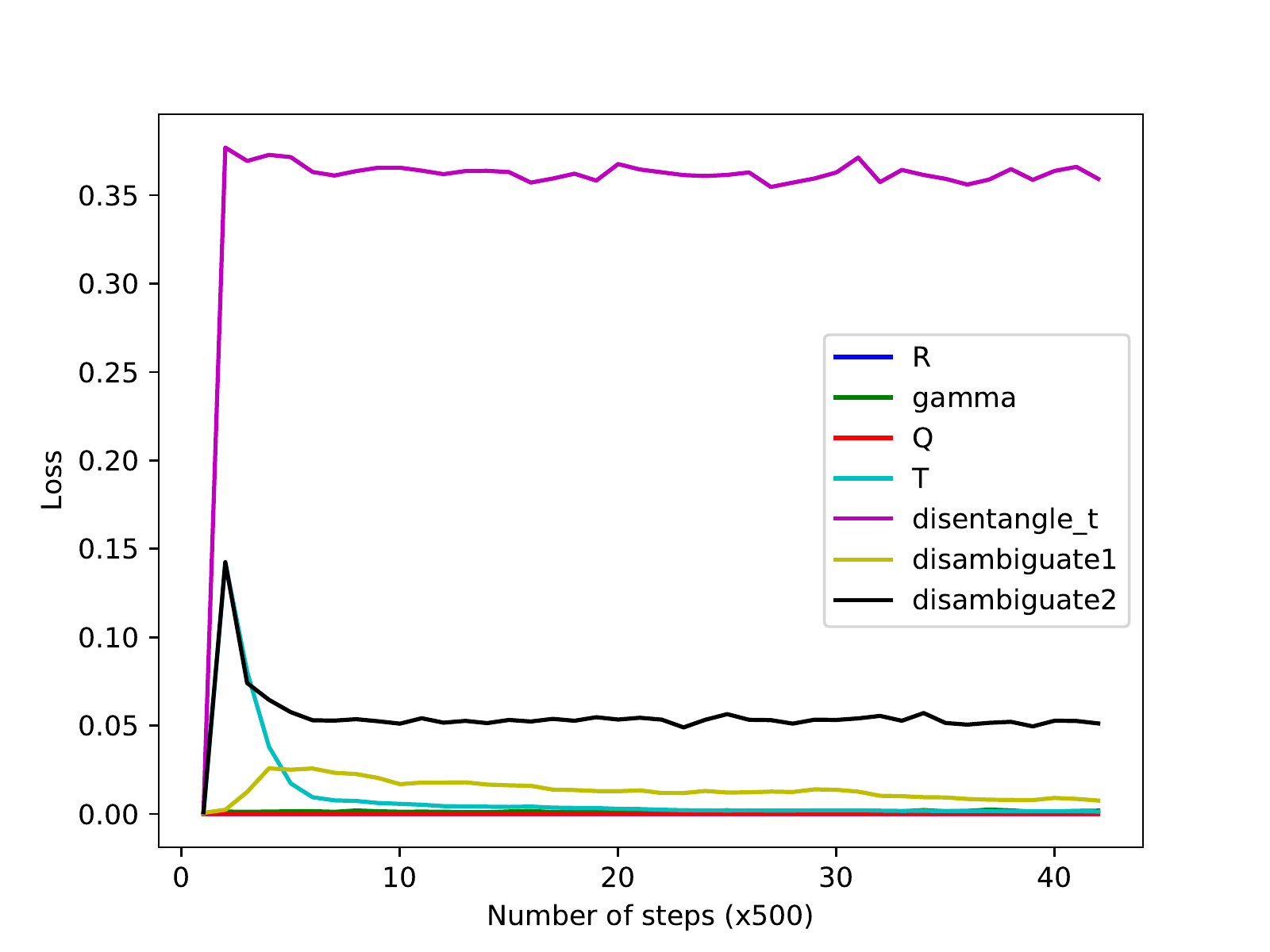}\hfill
    \includegraphics[width=.5\textwidth]{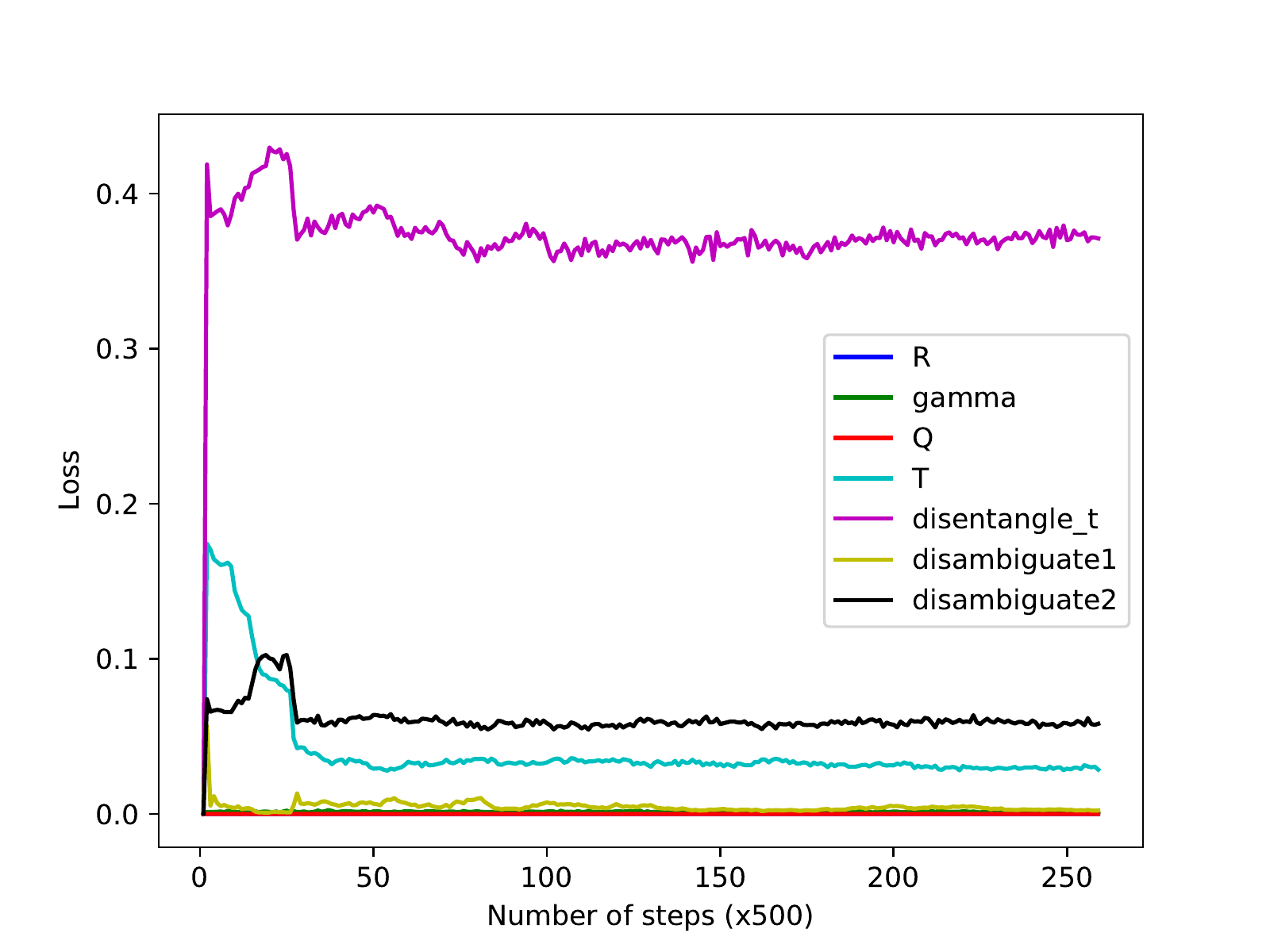}
    \caption{Visualisation of the losses over time. The left image shows the losses of the base agent, and the right image shows the losses of the failed fine-tuned agent in the target domain. Notice the difference in the transition loss }\label{low_losses}
\end{figure}

From Figure \ref{low_losses} we can observe that the losses are quite similar between the failed transfer and the converged base model (only the transition loss is slightly bigger). This confirms that although the embedding space is swapped, the losses are still small. Also note that the reward (R), gamma (discount) and Q (model free) losses are all 0 because there is no reward and no terminal state.

\subsection{Stuck local minimum.}
Now that it is clear why they often end up in the local minimum with switched sides in the embedding space, we can focus on why they keep stuck in this local optimum. We argue that a combination of two problems cause it to get stuck in the local optimum, namely 1) small losses for the initial weight initialisation in combination with a steep slope between this point and the global optimum and 2) lesser degree of freedom in the optimization process due to the frozen dynamics model. In the following paragraphs these will be explained.

\begin{figure}[htbp]
    \centering
    \includegraphics[width=0.62\textwidth]{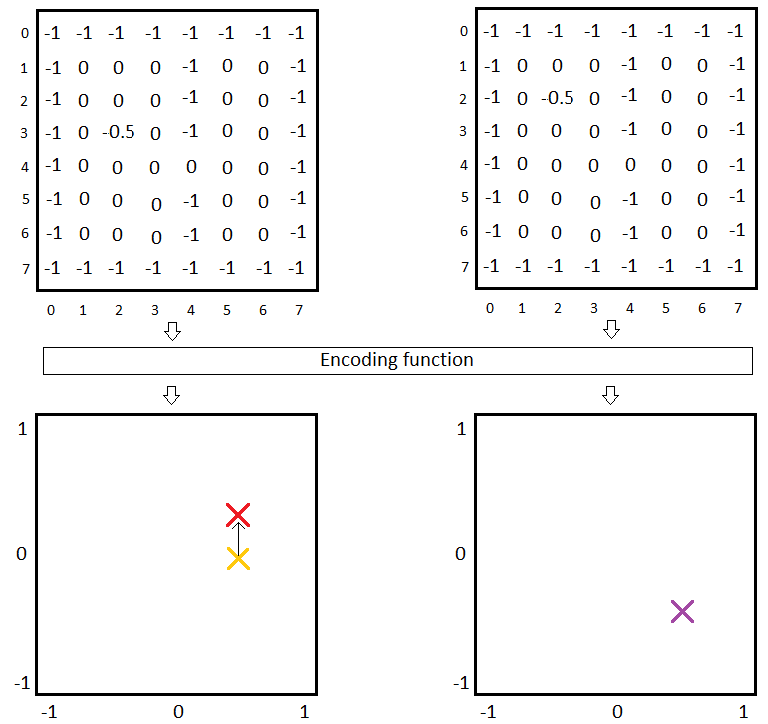}
    \caption{The first row contains the matrices with the actual input values, for the original player location (left) and for the player location when taken the up action (right). The yellow cross indicates the embedding value generated for the state with the original player location, red indicates the embedding value after applying the transition function when given the embedding value of the yellow cross and the up action, and the purple cross indicates the embedding value of the actual expected state. Note that the transition loss is the distance between the purple and red cross.}\label{lowtransition}
\end{figure}

First of all, let's take a more in-depth look why the losses are so small. We can feed the state of the inverse environment through the simple encoding function and then use the resulting embedding and an action to feed it through the transition function. We can then calculate the transition loss based on this new state embedding and the embedding of the actual expected state. Remember that the transition loss is simply the difference between the predicted embedding based on the state and an action, and the embedding of the actual state where you would end up in given the action. In Figure \ref{lowtransition} we can observe the end result of this process. 

We take the same player location (2,3) as before, thus the base model embedding is (-0.5,0) and the inverse embedding is (0.5,0) (illustrated with the yellow cross). Which if we wouldn't use rounding we find the actual corresponding inverse embedding (0.5, -0.25), as can be seen in the Figure \ref{low_comparison_embed}. When we apply the transition function (given by the lines in Figure \ref{low_comparison_embed}) to this embedding and the action 'up', we end up in the embedding value (0.5,0.25). Whereas if we actually get the inverse embedding of the state with a player location of (1,3), we first observe the base embedding: (-0.5, 0.5), then determine the simple encoding function for this state -0.5=0.5*-1, 0.5=0.5, and then finally calculate the inverse embedding by feeding it through the simple encoding function: 0.5=-0.5*-1, -0.5=-0.5, resulting in embedding value: (0.5, -0.5). Now we can approximate the transition loss by determining the distance between (0.5,0.25) and (0.5, -0.5). We can observe that this is quite small, and this is  approximately the maximum transition loss that will be encountered, since it will never be more than two edges away (at the sides of the environment its even only one edge away). 

We conclude that because of the switched sides the structure in the positions of the embedding values is maintained, and therefore the transition function will predict the embedding value of the opposite neighbour, resulting in a loss of maximum of two edge-lengths away (which for an embedding space with range -1 <> 1 and 6 positions is on average 0.33 * 2). However many embedding states (28 out of 64) reside near a side; which only have a loss of one edge length. Regarding all the other losses: these will be same as the base model since the positions of the states are very similar and the other losses mainly look at the distances between the states. 

It is easy to see that this small loss will result in a low point in the loss landscape. However a low point does not directly mean a local optimum, for example there can still lead a path downwards from this low point directly to the global optimum (lowest point). However we informally argue that this is not the case, and that there possible is quite a steep slope between the global minimum and the local minimum. So given 1) the normal environment and the converged base model and 2) the inverse environment and a failed fine-tuned model, we just showed that the embedding values are inversed for the same hidden state, for this we also showed that the losses are quite small because the structure of the values is maintained. Therefore the global optimum is at embedding value x and the (suboptimal inversed) local minimum is at value -x, and thus an embedding value $i \in \mathbb{Z}: -x < i < x $ will result in bigger losses because the structure is no longer maintained. More concretely, if we consider for a hidden state the embedding value of the base model $E_{b}$ and the embedding of the failed transfer $E_{t}$ a pair $(E_{b},E_{t})$, then it is easy to see that for all pairs the middle value of the two values of a pair is always 0 ($E_{b}=x, E_{t}=-x, 0=x-x$). Then if all embedding states need to converge past value 0 (which is the direct path between the stuck local minimum -x and the global optimum x) it is easy to see that for example the Disambiguate1 and the Disentangle loss will be at the highest point.

The second problem arises by the fact that the frozen dynamics model offers less freedom in the optimization process. We argue that the lack of freedom in combination with the first problem of getting caught in the local minimum will cause it to be stuck. We first describe this second problem in an intuitive manner, after which we provide it in a formal proof format. If we consider that the encoder and the dynamics weights have two separate loss landscapes. Then when the dynamics model weights change (e.g. get updated), then for a given input the loss changes (and its gradients w.r.t. the weights), which means that the loss landscape of the encoder model changes since the loss gets backpropagated through the dynamics model to the encoder(and even if the loss doesn't change but the weights do, the backpropagated gradients can change), see Theorem and Proof 2. Now if we consider a scenario where the encoder weights are stuck in a local minimum (which can happen because the loss function is non-convex), we can imagine that the dynamics model will keep trying to converge to an optimum solution which keeps changing the loss landscape of the encoder model which might help in escaping the local minimum, either by 1) making the local minimum of the encoder model a global minimum by changing the dynamics so that the model as a whole performs optimal, or 2) that by chance a random change in the encoder loss landscape opens a path out of the local minima towards the global optimum. However, when we freeze the dynamics model we lose this degree of freedom, and this in combination with the poor encoder weight initialisation, described before, with small but suboptimal losses causes it to get stuck.

\paragraph{theorem}
Given a model $\hat{y}$ which consist of an encoder and a dynamics model where the loss of the output of this model will be backpropagated through the dynamics model into the encoder. Then the changes in the dynamics model will change the loss landscape of the encoder model (this of course could be generalized to any model architecture with two sequential models with a shared loss). 
\paragraph{proof}
Given our model $\hat{y_{1}}=w_{1}x+b_{1}$ where $w_{1}$ and $b_{1}$ contain all our weights and biases of our encoder and dynamics model. To simplify the proof we use a simple squared error as our loss function (instead of the CRAR losses): $L_{1}=(y-\hat{y_{1}})^2$ which is the same as $L_{1}=(w_{1}x+b_{1}-y)^2$. Where $y$ is the target value. Then given a small change in the weights (e.g. after an update step) of our dynamics model which gives a different model $\hat{y_{2}}=w_{2}x+b_{2}$, where $\hat{y_{1}} \neq \hat{y_{2}}$ because $w_{1} \neq w_{2}$. Then given that the output of the models is different, it follows that the loss for a given input (x) is different between the two models, because: $(\hat{y_{1}}-y)^2 \neq (\hat{y_{2}}-y)^2$, thus $L_{1} \neq L_{2}$. Then it also follows that the gradients of the losses with respect to the weights($\frac{\partial L_{1}}{\partial w_{1}} = 2x(w_{1}x+b_{1}-y)$) are different: $\frac{\partial L_{1}}{\partial w_{1}} \neq \frac{\partial L_{2}}{\partial w_{2}}$ because $2x(w_{1}x+b_{1}-y) \neq 2x(w_{2}x+b_{2}-y)$ since $w_{1} \neq w_{2}$.
Given that the encoder is part of the model and that the losses and gradients are different between the models, we can conclude that the loss landscape of the encoder model changes as the dynamics model changes its weights. Which implies that if we freeze the dynamics model weights, the encoder loss landscape will not change by means of changes in the dynamics model.

\begin{wrapfigure}{r}{0.5\textwidth}
      \centering
      \includegraphics[scale=0.65]{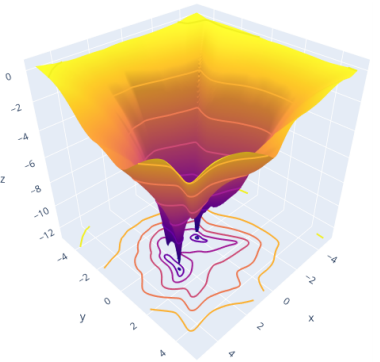}
      \caption{Visualisation of a loss landscape of a neural network\cite{DLtheory}.}\label{losslandscape}
      \vspace{-25pt}
\end{wrapfigure}

To future illustrate this reduced freedom in the optimization process, consider the loss landscape of Figure \ref{losslandscape}, and take for x the set of weights of the encoder model, for y the the set of weights of the dynamics model, and for z the loss function (w.r.t. the weights of the encoder and dynamics model). Which of course is an oversimplification since the plot is displaying the y- and x-axis as an numerical variable, instead of a set of numerical values, but for this point it suits the needs as we can assume that both models only contain a single weight. Using this visualisation it is easy to see that if we freeze the dynamics model we can remove the y-axis since this is now a single constant value instead of a possible range of values, meaning that we reduce the degree of freedom of the optimization process by one dimension.

We conclude that the combination of the two problems cause it to get stuck in a local optimum: the weight initialisation causes it to immediately find the swapped-sides-embedding solution with small losses and a steep slope between this solution and the global optimum, thus finding it self in a local minimum (problem 1), and the dynamics model is not able to alleviate this problem by changing the loss landscape because it is frozen (problem 2). Further concluding that freezing the dynamics model can result in a sub optimal solution, worse than the pre-trained model or training from scratch. The sub optimal performance is caused by the incompatible embedding space and thus poor task performance. We argue that due to randomness in the learning process sometimes the transfer escapes the local optimum and thus succeeds, and sometimes it gets stuck and thus fails. 

\section{Losses of high-dimensional transfers}
\begin{figure}[htbp]
    \includegraphics[width=.33\textwidth]{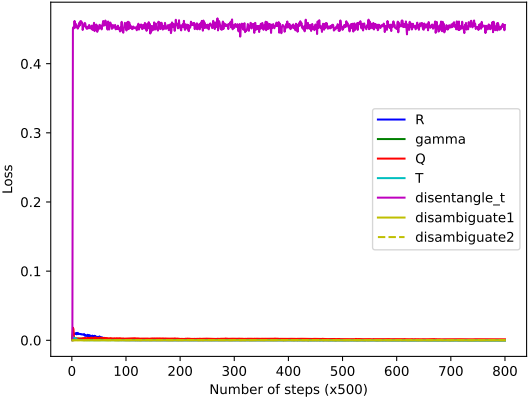}\hfill
    \includegraphics[width=.33\textwidth]{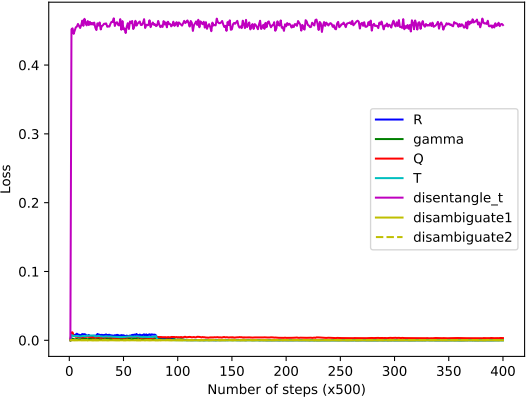}\hfill
    \includegraphics[width=.33\textwidth]{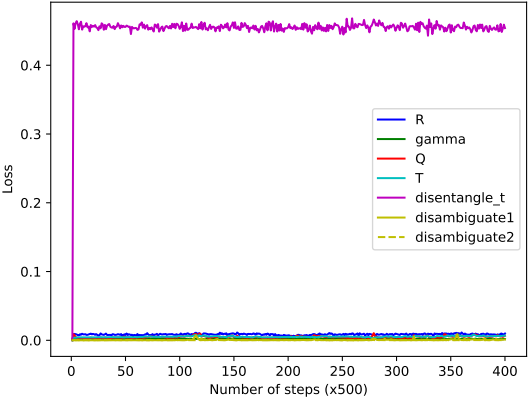}
    \caption{Visualisation of the losses associated to (from left to right): a base model, a successful transfer, and a failing transfer.}\label{highdim_perforamncelosses}
\end{figure}

As before we find that the losses are quite small for the failing transfers (Figure \ref{highdim_perforamncelosses}). However, it can be noted that the reward, transition and q losses remain slightly higher. This indicates that the model did not converge to the same global optimum, and thus did not converge to the same embedding space. Remarkable here is that the losses are very small, but the agent is not able to perform the task at all (score of -5). This phenomena is caused by that the agent doesn't explicitly optimize the reward, it optimizes the implicit losses. This means that even if the agent is able to accurately estimate the reward for a state action pair, then the model could still have trouble collecting the rewards: for instance with the transition model being stuck in the swapped sides local minimum.


\section{Study of Approach Variations}
\begin{figure}[htbp]
  \includegraphics[width=0.55\linewidth]{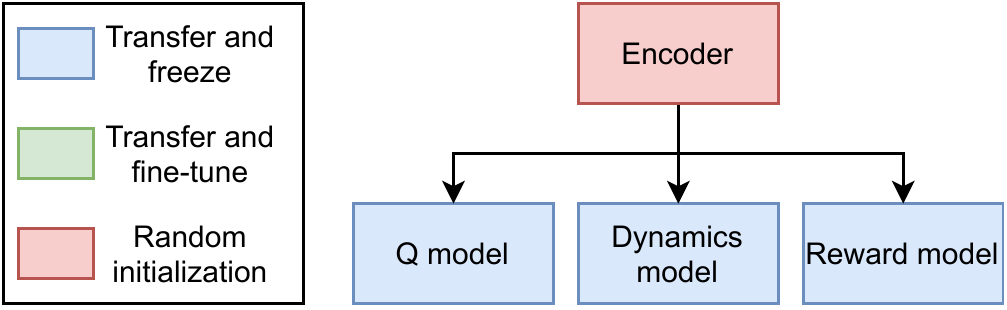}
  \centering
  \caption{Variation 1: re-learn the encoder from scratch}\label{var1}
\end{figure}

\section{Study of learning rate}
In the final experiment we experiment with different learning rates for the transfers (using the original proposed approach). We use the following learning rates: 0.00001, 0.00005, 0.0001, 0.0005, 0.001 and  0.005. To exclude the effect of different starting models, we use a single base model: basemodel2. Using this single base model, we will execute six transfers per learning rate, so a total of 36 transfers.

\begin{figure}[htbp]
    \includegraphics[width=.5\textwidth]{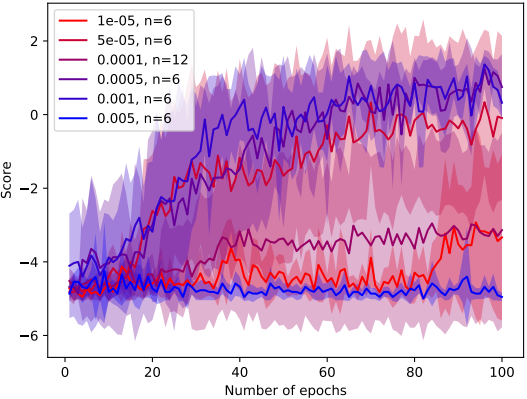}\hfill
    \includegraphics[width=.5\textwidth]{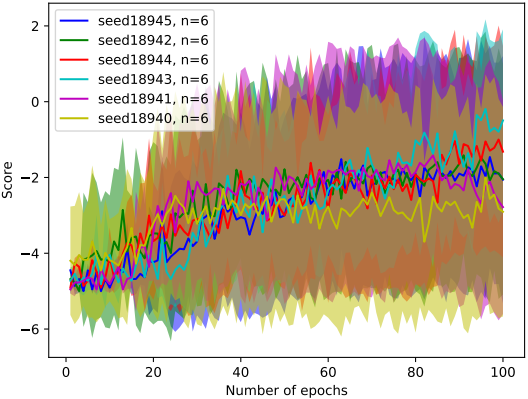}
    \caption{Visualisation of the average and the standard deviation of the score of 36 transfers, with the same base model but six different learning rates with six seeds each. In the left plot the data is grouped by the learning rate, and in the right group the data is grouped by the seed. The six extra samples for learning rate 0.0001 have been obtained using six different seeds, these have been omitted from the right plot due to irrelevance. }\label{highidim_lr}
\end{figure}

From Figure \ref{highidim_lr} we can observe that 0.001, 0.0005 and 0.00005 mostly have successful transfers, with 0.00005 having the biggest variance. We find that 0.005, 0.0001 and 0.00001 mostly fail, but 0.0001 and 0.00001 have quite a big variance. We conclude that the main trend we can observe from this experiment is that in the middle of the learning rates they perform quite well, but to the more extremes: 0.005 and 0.00001 they fail, which is as expected. The only exception being 0.0001, which appears to perform worse than both its neighbouring learning rates, even after increasing its sample size. We attribute this strange occurrence to the small sample sizes (e.g. 5e-05 might appear to perform much better than it actually should). Therefore the experiment results provide a general understanding of the behaviour of the different learning rates, but learning rate specific performance should be taken with a grain of salt. 

\section{Hyperparameters and compute resources per experiment}
For each experiment the used computing resources and the total usage time of the compute resource for the experiment are given:
\begin{itemize}
  \item Low dimensional maze: Nvidia GTX 1050Ti for 10 days. 
  \item High dimensional maze: 
  \begin{itemize}
      \item Original approach with 5 different base models (section 5.2.1): 4 nodes with an Intel® Xeon® Bronze 3104 with 256GB memory and 4xNvidia 1080Ti for 2 days
      \item Visualisation of embedding per base model (section 5.2.2): Nvidia GTX 1050Ti for 30 minutes
      \item Performance of different approach (section 5.2.3): 4 nodes with an Intel® Xeon® Bronze 3104 with 256GB memory and 4xNvidia 1080Ti for 2 days
    \end{itemize}
\end{itemize}

\end{document}